\documentclass[11pt, a4paper]{lumia}

\usepackage[sort&compress]{natbib}
\usepackage{xspace}

\theoremstyle{plain}

\newtheorem*{proposition*}{Proposition}

\theoremstyle{definition}

\theoremstyle{definition}

\def\eqref#1{equation~\ref{#1}}

\usepackage{graphicx}           
\usepackage{tikz}               
\usepackage[edges]{forest}      

\usepackage{url}                
\usepackage{xurl}               

\usepackage{array}              
\usepackage{longtable}          
\usepackage{multirow}           
\usepackage{makecell}           
\usepackage{ragged2e}           

\usepackage{mathtools}          
\usepackage{nicefrac}           

\usepackage{algorithm}          
\usepackage{algorithmicx}       
\usepackage{algpseudocode}      
\usepackage{listings}           

\usepackage{subcaption}         
\usepackage{wrapfig}            
\usepackage{placeins}           
\usepackage[export]{adjustbox}  

\usepackage{xspace}             
\usepackage[normalem]{ulem}     
\usepackage{CJKutf8}            

\usepackage[tikz]{bclogo}       
\usepackage[framemethod=tikz]{mdframed} 

\usepackage{lipsum}             
\usepackage{tocloft}            
\usepackage{afterpage}          
\usepackage{bbding}             
\usepackage{epigraph}           
\usepackage{minitoc}            
\usepackage{multicol}           
\usepackage{textgreek}          

\newcolumntype{P}[1]{>{\RaggedRight\arraybackslash}p{#1}}

\definecolor{uclablue}{RGB}{39, 116, 174}
\definecolor{bigaired}{RGB}{156, 0, 0}
\definecolor{myblue}{HTML}{598BE7}
\definecolor{mildblue}{RGB}{31,119,180}
\definecolor{sectionblue}{RGB}{70, 130, 180}
\definecolor{methodblue}{RGB}{0, 150, 136}
\definecolor{bgblue}{RGB}{245,243,253}
\definecolor{ttblue}{RGB}{91,194,224}
\definecolor{mygreen}{rgb}{0.64, 0.56, 0.88}
\definecolor{myyellow}{rgb}{0.68, 0.6, 0.1}
\definecolor{fancygreen}{rgb}{0.33, 0.68, 0.20}
\definecolor{salmon}{rgb}{0.94, 0.52, 0.49}
\definecolor{tablegreen}{rgb}{0.82, 0.94, 0.75}
\definecolor{tableblue}{rgb}{0.81, 0.90, 0.94}
\definecolor{tablered}{rgb}{0.97, 0.85, 0.85}
\definecolor{tableorange}{rgb}{0.96, 0.85, 0.81}
\definecolor{myorange}{rgb}{1.0, 0.49, 0.0}
\definecolor{casepurple}{RGB}{112,96,200}
\definecolor{tlgreen}{rgb}{0.33, 0.68, 0.20}
\definecolor{darkgreen}{RGB}{0,100,0}
\definecolor{darkred}{RGB}{200, 0, 0}
\definecolor{customyellow}{HTML}{FFFACD}
\definecolor{refinegreen}{RGB}{0, 128, 75}
\definecolor{scoregreen}{RGB}{34, 139, 34}
\definecolor{hidden-blue}{RGB}{194,232,247}
\definecolor{hidden-black}{RGB}{20,68,106}
\definecolor{yes}{HTML}{C6EFCE}
\definecolor{no}{HTML}{FFC7CE}
\definecolor{partial}{HTML}{FFEB9C}
\definecolor{external}{HTML}{D9E1F2}
\definecolor{hdr}{HTML}{F2F2F2}
\definecolor{GRPOrow}{gray}{0.96}
\definecolor{FlowRLrow}{RGB}{225,236,255}
\definecolor{FlowBlue}{RGB}{80,120,210}
\definecolor{GRPOGray}{gray}{0.35}

\hypersetup{
    colorlinks=true, 
    citecolor=uclablue, 
    linkcolor=bigaired,
    urlcolor=darkblue
}

\setlist[itemize]{leftmargin=20pt, noitemsep, topsep=0pt}

\NewDocumentCommand{\kaiyan}{mO{}}{\textcolor{purple}{\textsuperscript{\textit{kaiyan}}\textsf{\textbf{\small[#1]}}}}
\NewDocumentCommand{\yuxin}{mO{}}{\textcolor{cyan}{\textsuperscript{\textit{yuxin}}\textsf{\textbf{\small[#1]}}}}
\NewDocumentCommand{\bx}{mO{}}{\textcolor{green}{\textsuperscript{\textit{bx}}\textsf{\textbf{\small[#1]}}}}
\NewDocumentCommand{\at}{mO{}}{\textcolor{red}{\textsuperscript{\textit{AT}}\textsf{\textbf{\small[#1]}}}}
\NewDocumentCommand{\re}{mO{}}{\textcolor{blue}{\textsuperscript{\textit{RE}}\textsf{\textbf{\small[#1]}}}}
\NewDocumentCommand{\ybsun}{mO{}}{\textcolor{magenta}{\textsuperscript{\textit{youbang}}\textsf{\textbf{\small[#1]}}}}
\NewDocumentCommand{\runze}{mO{}}{\textcolor{orange}{\textsuperscript{\textit{runze}}\textsf{\textbf{\small[#1]}}}}
\NewDocumentCommand{\add}{mO{}}{\textcolor{darkgreen}{\textsuperscript{\textit{Maybe Consider Discuss}}\textsf{\textbf{[#1]}}}}

\newcommand{\cmark}{\textcolor{darkgreen}{\boldmath$\checkmark$}}
\newcommand{\xmark}{\textcolor{darkred}{\boldmath$\times$}}

\newenvironment{itemize*}%
 {\leftmargini=10pt\begin{itemize}%
  \setlength{\itemsep}{0pt}%
  \setlength{\parskip}{0pt}%
  }%
 {\end{itemize}}

\newenvironment{enumerate*}%
 {\begin{enumerate}%
  \setlength{\itemsep}{0pt}%
  \setlength{\parskip}{0pt}}%
 {\end{enumerate}}

\newcommand{\cellstatus}[1]{%
  \begingroup
  \StrTrim{#1}[\statusval]%
  \IfStrEq{\statusval}{Yes}{\cellcolor{yes}\cmark}{}%
  \IfStrEq{\statusval}{No}{\cellcolor{no}\xmark}{}%
  \IfBeginWith{\statusval}{Yes (}{\cellcolor{yes}\cmark~\textit{\statusval\unskip}}{}%
  \IfStrEq{\statusval}{Partial}{\cellcolor{partial}\textbf{Partial}}{}%
  \IfStrEq{\statusval}{External}{\cellcolor{external}\textbf{External}}{}%
  \endgroup
}

\newtcolorbox{myboxi}[1][]{
  breakable,
  title=#1,
  colback=red!5,
  colbacktitle=red!5,
  coltitle=black,
  fonttitle=\bfseries,
  bottomrule=0pt,
  toprule=0pt,
  leftrule=2pt,
  rightrule=2pt,
  titlerule=0pt,
  arc=0pt,
  outer arc=0pt,
  colframe=red,
}

\newtcolorbox{myboxnote}[1][]{
  breakable,
  title=#1,
  colback=orange!0,
  colbacktitle=orange!0,
  coltitle=black,
  fonttitle=\bfseries,
  bottomrule=0pt,
  toprule=0pt,
  leftrule=2pt,
  rightrule=2pt,
  titlerule=0pt,
  arc=0pt,
  outer arc=0pt,
  colframe=orange,
}

\newtcolorbox{myboxii}[1][]{
  breakable,
  freelance,
  title=#1,
  colback=white,
  colbacktitle=white,
  coltitle=black,
  fonttitle=\bfseries,
  bottomrule=0pt,
  boxrule=0pt,
  colframe=white,
  overlay unbroken and first={
  \draw[red!75!black,line width=3pt]
    ([xshift=5pt]frame.north west) -- 
    (frame.north west) -- 
    (frame.south west);
  \draw[red!75!black,line width=3pt]
    ([xshift=-5pt]frame.north east) -- 
    (frame.north east) -- 
    (frame.south east);
  },
  overlay unbroken app={
  \draw[red!75!black,line width=3pt,line cap=rect]
    (frame.south west) -- 
    ([xshift=5pt]frame.south west);
  \draw[red!75!black,line width=3pt,line cap=rect]
    (frame.south east) -- 
    ([xshift=-5pt]frame.south east);
  },
  overlay middle and last={
  \draw[red!75!black,line width=3pt]
    (frame.north west) -- 
    (frame.south west);
  \draw[red!75!black,line width=3pt]
    (frame.north east) -- 
    (frame.south east);
  },
  overlay last app={
  \draw[red!75!black,line width=3pt,line cap=rect]
    (frame.south west) --
    ([xshift=5pt]frame.south west);
  \draw[red!75!black,line width=3pt,line cap=rect]
    (frame.south east) --
    ([xshift=-5pt]frame.south east);
  },
}

\tcbset{
  takeawaysbox/.style={
    title=Takeaways,
    colback=lightblue!80,
    colframe=black,
    fonttitle=\bfseries\small,
    coltitle=white,
    colbacktitle=black,
    enhanced,
    attach boxed title to top left={xshift=2.5mm,yshift=-2.5mm},
    boxed title style={rounded corners, size=small, colframe=black, colback=black},
    width=\linewidth,
    arc=3.5mm
  }
}

\mdfdefinestyle{mystyle}{%
  rightline=true,
  innerleftmargin=10,
  innerrightmargin=10,
  outerlinewidth=3pt,
  topline=false,
  rightline=true,
  bottomline=false,
  skipabove=\topsep,
  skipbelow=\topsep
}

\tikzset{%
    every node/.style={font=\tiny},
    parent/.style =          {align=center,text width=2cm,rounded corners=3pt, line width=0.3mm, fill=gray!10,draw=gray!80},
    child/.style =           {align=center,text width=2.0cm,rounded corners=3pt, fill=blue!10,draw=blue!80,line width=0.3mm},
    grandchild/.style =      {align=center,text width=2cm,rounded corners=3pt},
    greatgrandchild/.style = {align=center,text width=1.5cm,rounded corners=3pt},
    greatgrandchild2/.style = {align=center,text width=1.5cm,rounded corners=3pt},    
    referenceblock/.style =  {align=center,text width=1.5cm,rounded corners=2pt},
    pretrain/.style =           {align=center,text width=2.0cm,rounded corners=3pt, fill=blue!10,draw=blue!80,line width=0.3mm},   
    pretrain_work/.style =           {align=center, text width=8.5cm,rounded corners=3pt, fill=blue!10,draw=blue!0,line width=0.3mm},  
    template/.style =           {align=center,text width=2.0cm,rounded corners=3pt, fill=red!10,draw=red!80,line width=0.3mm},   
    template_work/.style =           {align=center,text width=8.5cm,rounded corners=3pt, fill=red!10,draw=red!0,line width=0.3mm},    
    answer/.style =           {align=center,text width=2.0cm,rounded corners=3pt, fill= cyan!10,draw= cyan!80,line width=0.3mm},   
    answer_work/.style =           {align=center,text width=8.5cm,rounded corners=3pt, fill= cyan!10,draw= cyan!0,line width=0.3mm},      
    multiple/.style =           {align=center,text width=2.0cm,rounded corners=3pt, fill= orange!10,draw= orange!80,line width=0.3mm},   
    multiple_work/.style =           {align=center,text width=8.5cm,rounded corners=3pt, fill= orange!10,draw= orange!0,line width=0.3mm},        
    tuning/.style =           {align=center,text width=2.0cm,rounded corners=3pt, fill= magenta!10,draw= magenta!80,line width=0.3mm},   
    tuning_work/.style =           {align=center,text width=8.5cm,rounded corners=3pt, fill= magenta!10,draw= magenta!0,line width=0.3mm},          
}

\newcommand{\lstbg}[3][0pt]{{\fboxsep#1\colorbox{#2}{\strut #3}}}

\lstdefinelanguage{diff}{
  basicstyle=\ttfamily\small,
  morecomment=[f][\lstbg{red!20}]-,
  morecomment=[f][\lstbg{green!20}]+,
}

\lstdefinelanguage{diffpython}{
  language=diff,
  morekeywords={def, if, else, for, while, return, import, from, as, class, with, try, except, finally, raise, lambda, and, or, not, in, is, None, True, False},
  morecomment=[l]{\#},
  morestring=[b]",
  morestring=[b]',
}

\definecolor{gaincolor}{RGB}{0,128,0}
\definecolor{dropcolor}{RGB}{190,40,40}
\definecolor{neutralcolor}{RGB}{110,110,110}
\newcommand{\gain}[1]{\textcolor{gaincolor}{#1}}
\newcommand{\drop}[1]{\textcolor{dropcolor}{#1}}
\newcommand{\neutral}[1]{\textcolor{neutralcolor}{#1}}
\newcommand{\scorecell}[2]{\begin{tabular}[t]{@{}l@{\hspace{0.18em}}l@{}}#1 & {\tiny #2}\end{tabular}}
\newcommand{\rawscorecell}[1]{\begin{tabular}[t]{@{}l@{\hspace{0.18em}}l@{}}#1 & {\tiny \phantom{+0.0}}\end{tabular}}
\newcommand{\avgcell}[2]{\begin{tabular}[t]{@{}l@{\hspace{0.18em}}l@{}}#1 & (#2)\end{tabular}}
\newcommand{\rawavgcell}[1]{\begin{tabular}[t]{@{}l@{\hspace{0.18em}}l@{}}#1 & \phantom{(+0.00)}\end{tabular}}
\newcommand{\headtworow}[1]{\raisebox{-1.5ex}[0pt][0pt]{#1}}

\setheadertext{LUMIA Lab}

\correspondingemail{\emailicon{} wangjiarui1@sjtu.edu.cn, lin.zhouhan@gmail.com \quad $^*$ Equal Contribution. \quad $^\ddagger$ Corresponding Author. \\
\faGithub\ \href{https://github.com/LUMIA-Group/MemSFT}{LUMIA-Group/MemSFT}
\quad
\faHuggingface\ \href{https://huggingface.co/collections/Jiarui-Wang/memsft}{Jiarui-Wang/MemSFT}}

\title{MemSFT: Mitigating Alignment Tax with an External Parametric Memory}
\setheadertitle{MemSFT: Mitigating Alignment Tax with an External Parametric Memory}

\author{%
  {\Authfont Jiarui Wang$^{1,2,3*}$, Xiang Shi$^{1*}$, Jiaqi Cao$^{1}$, Rubin Wei$^{1,2}$, Xiquan Wang$^{1,2}$, Hao Sun$^{1,2}$}\\
  {\Authfont Jingzhi Wang$^{1}$, Zhiqi Yang$^{1}$, Qipeng Guo$^{2}$, Bowen Zhou$^{2,4}$, Zhouhan Lin$^{1,2\ddagger}$}\\
  $^1$ LUMIA Lab, School of Artificial Intelligence, Shanghai Jiao Tong University\\
  $^2$ Shanghai AI Laboratory \quad $^3$ School of Computer Science, Shanghai Jiao Tong University\\
  $^4$ Tsinghua University
}

\begin{document}

\begin{abstract}
Adapting Large Language Models (LLMs) to specialized domains often incurs an alignment tax, as fine-tuning on domain-specific tasks can cause catastrophic forgetting and substantially degrade performance on general tasks.
We propose \textbf{MemSFT}, which mitigates the alignment tax by decoupling domain specialization from backbone parameter updates through a plug-and-play parametric memory.
The memory is trained to imitate the behavior of a non-parametric retriever operating over domain data, thereby memorizing knowledge and patterns that would otherwise be accessed through retrieval.
Once trained on a specific domain, the memory can be reused across LLMs of different sizes.
During generation, a learned router dynamically fuses the output distributions of the memory and backbone at each decoding step, allowing domain expertise to be invoked selectively.
Across biology, geoscience, and law, evaluations with models ranging from Qwen3-8B to Qwen3-235B-A22B show that MemSFT consistently improves domain performance with negligible degradation in general performance, whereas full SFT suffers severe forgetting on general tasks.
Overall, our results demonstrate a practical path to decoupling general model capabilities from domain-specific knowledge at the parameter level, thereby equipping LLMs with new specialized capabilities without compromising their general capabilities.

\end{abstract}

\maketitle

\section{Introduction}
Large Language Models (LLMs) have emerged as powerful foundation models, exhibiting strong few-shot transfer, instruction following, and open-ended generation across a wide range of tasks \citep{brown2020language,bommasani2021opportunities,ouyang2022training}. However, deploying LLMs in specialized domains often requires expertise and reliability beyond generic assistant capabilities. This gap is particularly acute in scientific and professional domains such as biology, geoscience, and law, where models must handle symbolic, numerical, and structured prediction tasks that modern LLMs do not reliably solve \citep{singhal2025toward,he2024biology,liu2025openswi,yue2024lawllm}. A common response is supervised fine-tuning (SFT), which has proven effective for improving task following and domain-specific performance \citep{wei2021finetuned,chung2024scaling,ouyang2022training}. In practice, domain SFT is usually applied on top of post-trained assistant models, after general instruction tuning or agent-oriented post-training has established broad instruction-following and reasoning abilities.

\begin{figure*}[t]
    \centering
    \includegraphics[width=\textwidth]{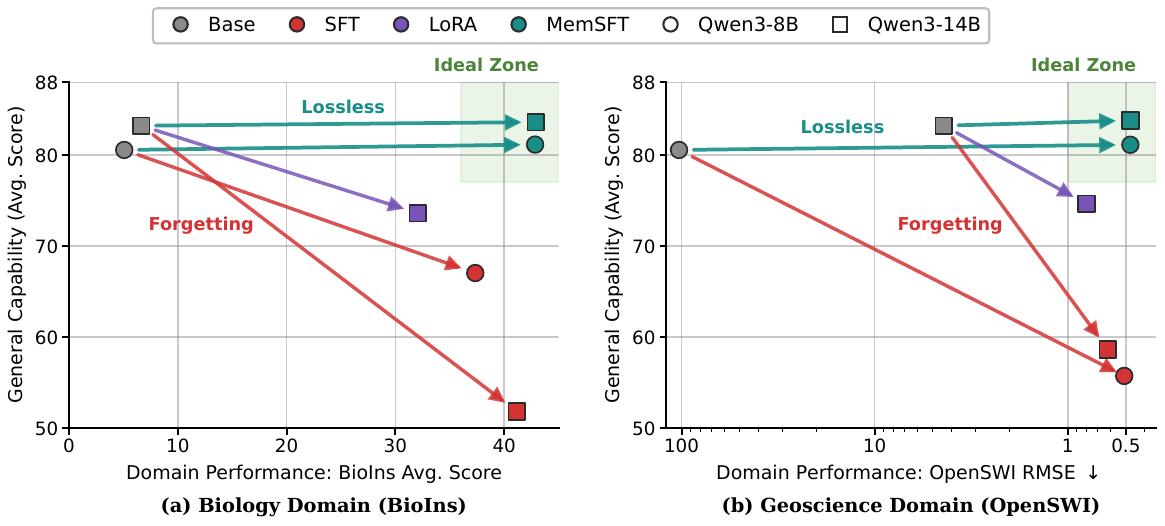}
    \caption{Domain performance and general capability retention across two scientific domains: Biology-Instructions (BioIns) for biology and OpenSWI for geoscience, evaluated with Qwen3 backbones. Within each domain, MemSFT reuses the same domain-specific 8B memory across different backbone sizes. SFT and LoRA increase specialization but reduce general capability, whereas MemSFT approaches the ideal zone while preserving general capability. The OpenSWI x-axis uses a logarithmic scale. Qwen3-32B, Qwen3-235B-A22B, and full numerical results are reported in Tables~\ref{tab:bioins-main} and~\ref{tab:openswi-main}.}
    \label{fig:main-effect}
\end{figure*}

However, the parameter updates that make SFT effective also expose LLMs to \textbf{catastrophic forgetting}, which has long been studied as the loss of previously acquired capabilities when a model is further optimized for a new task, domain, or distribution \citep{mccloskey1989catastrophic,french1999catastrophic,parisi2019continual}. Recent analyses of LLM fine-tuning show that such forgetting is not merely a legacy issue of smaller neural models, but can be especially severe when adapting modern post-trained LLMs \citep{luo2025empirical,li2024revisiting,shi2025continual,zheng2025towards,dong2024abilities}. Notably, \citet{ji2025language} demonstrate that post-alignment language models exhibit an ``elastic'' nature; subsequent fine-tuning can disrupt their alignment equilibrium and rapidly shift model behavior back toward the broader distribution shaped by pre-training, eroding post-training capabilities such as instruction following, reasoning, and aligned response behavior. These findings show that domain SFT can incur a substantial alignment tax, with specialized gains coming at the expense of broadly useful post-training capabilities.

PEFT methods such as LoRA reduce forgetting by freezing the backbone and training only lightweight adapter modules \citep{hu2022lora,houlsby2019parameter}. However, recent studies reveal that PEFT methods still suffer from systematic catastrophic forgetting, exhibiting a strong inverse linear relationship between target task performance and the retention of general capabilities \citep{kalajdzievski2024scaling,biderman2024lora}. In essence, LoRA constrains parameter changes to a low-rank subspace but still leaves the model susceptible to catastrophic forgetting; it merely shifts the trade-off curve between domain-specific performance and the retention of general capabilities.
 
To move beyond this trade-off, we turn to a different architectural paradigm in which domain knowledge is stored in an external parametric memory while the backbone LLM remains intact. Memory Decoder \citep{cao2026memory} and MLP Memory \citep{wei2025mlp} instantiate this paradigm by distilling non-parametric retrieval behavior into trainable memory modules that can serve as an external memory to existing LLMs. These approaches demonstrate strong gains in pretraining-centric settings, including language modeling perplexities and knowledge-intensive evaluation. However, they primarily focus on pretrained language models and do not systematically examine the catastrophic forgetting that arises when post-trained instruct models are further specialized with domain data.

In this work, we propose \textbf{MemSFT}, which advances the external parametric-memory paradigm to domain specialization of modern large-scale post-trained LLMs up to 235B parameters. MemSFT mitigates the alignment tax by externalizing domain knowledge into a separate parametric memory while keeping the backbone LLM frozen. Following Memory Decoder \citep{cao2026memory} and MLP Memory \citep{wei2025mlp}, we train the memory to approximate the output distribution of a non-parametric retriever~\citep{khandelwal2019generalization}. Here, the retriever is constructed over target-domain SFT data, enabling the memory to internalize instruction-level domain expertise. During inference, the memory augments the frozen backbone by fusing their next-token distributions, with a learned router determining the memory's contribution at each decoding step. Once trained on a specific domain, the memory can be reused across LLMs of different sizes.

\Needspace{13\baselineskip}
Experimental results across biology, geoscience, and law, with Qwen3 backbones at multiple scales, demonstrate three key advantages of MemSFT:
\begin{itemize}
    \item MemSFT substantially improves domain performance across backbone LMs with negligible degradation in general capabilities, while SFT and LoRA suffer from severe forgetting (Figure~\ref{fig:main-effect}; Tables~\ref{tab:bioins-main}, \ref{tab:openswi-main}, and \ref{tab:lawbench}).
    
    \item For a given domain, the same memory can be reused as a plug-and-play module across Qwen3 backbones ranging from Qwen3-8B to Qwen3-235B-A22B, without retraining it for each backbone (Tables~\ref{tab:bioins-main} and~\ref{tab:openswi-main}).
    
    \item MemSFT requires far less adaptation compute: adapting all four Qwen3 backbones requires 9.23 EFLOPs, only 0.22x the 41.05 EFLOPs required by full SFT (Table~\ref{tab:adaptation-flops}).
\end{itemize}
 
\section{Method}
\label{sec:method}

\subsection{Overview}
We present \textbf{MemSFT}, a modular approach that decouples domain specialization from backbone parameter updates through a plug-and-play parametric memory. MemSFT has two core components: a specialized memory training procedure that aligns the memory LM with retrieval-based teacher distributions (Section~\ref{sec:datastore-construction} and Section~\ref{sec:training}), and a dynamic inference mechanism that enables the same memory to augment different backbone LLMs through token-level interpolation (Section~\ref{sec:inference}).

\begin{figure*}[t]
    \centering
    \includegraphics[width=\textwidth]{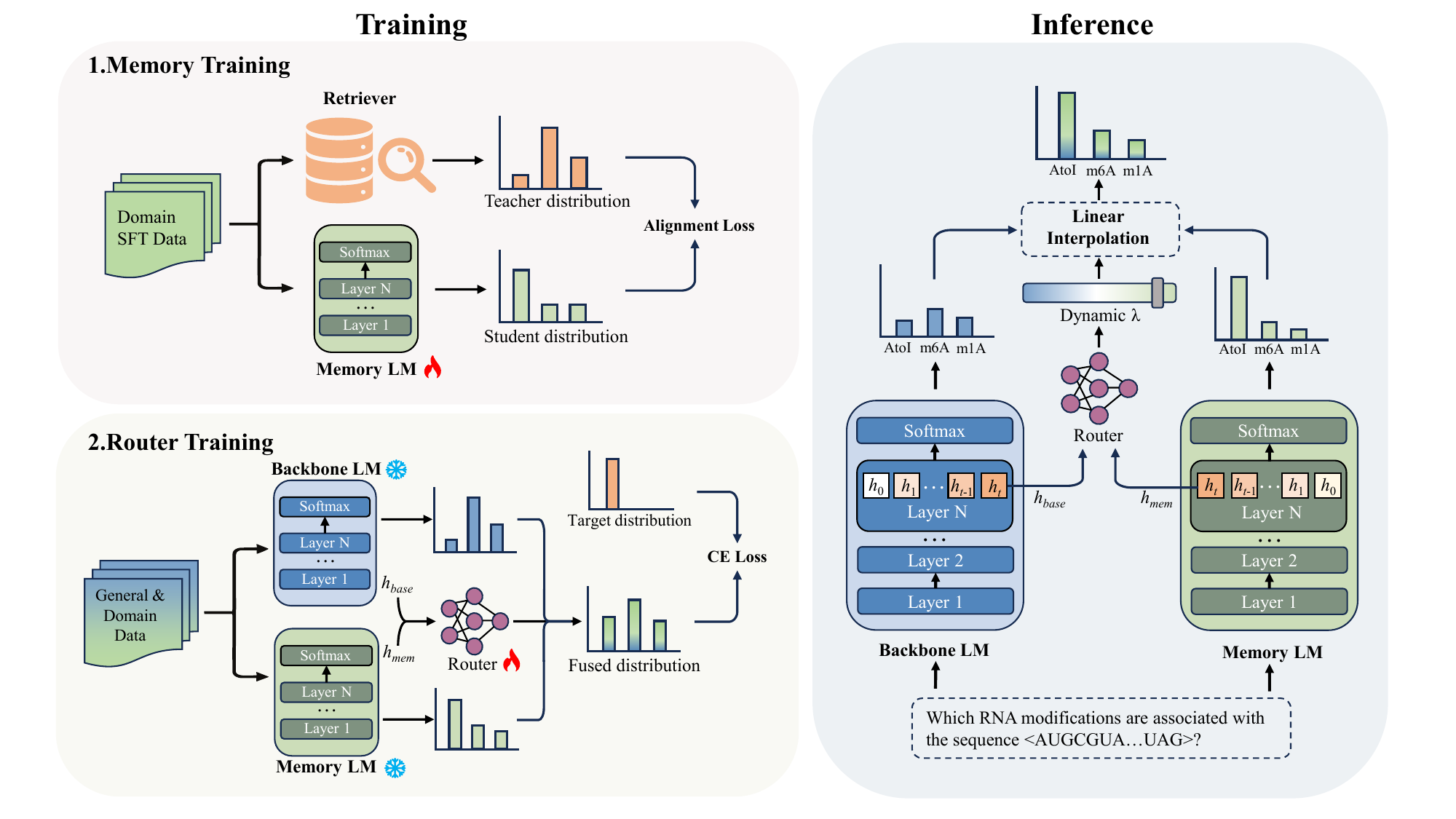}
    \caption{Overview of MemSFT architecture. \textbf{Upper-left}: \S~\ref{sec:datastore-construction} \& \S~\ref{sec:training}: During memory training, a QA-oriented datastore provides retrieval-based teacher distributions, and the memory LM learns to mimic them as a standalone parametric memory. \textbf{Lower-left}: \S~\ref{sec:training}: During router training, the backbone LM and memory LM are frozen, and a lightweight router learns token-wise interpolation weights. \textbf{Right}: \S~\ref{sec:inference}: During inference, the backbone and memory process the same input sequence in parallel, and their output distributions are fused for final prediction.}
    \label{fig:memsft-architecture}
\end{figure*}

As illustrated in Figure~\ref{fig:memsft-architecture}, MemSFT first trains the memory LM to approximate retrieval-based teacher distributions from a token-level datastore (upper-left part), then trains the router with both the backbone and memory frozen (lower-left part), and finally uses router-controlled fusion of the backbone and memory outputs to produce the next-token distribution (right part). 
 
\subsection{Constructing Retrieval-Based Supervision}
\label{sec:datastore-construction}

MemSFT first constructs a QA-oriented datastore to provide retrieval-based supervision for memory training. The datastore acts as a token-level RAG teacher: given a decoding context, it retrieves similar contexts from the domain SFT corpus and converts their corresponding next-token values into a retrieval-based teacher distribution \citep{khandelwal2019generalization}.
Let the domain SFT corpus be
\begin{equation}
\mathcal{D}_{\mathit{sft}} = \{(q^{(i)}, a^{(i)})\}_{i=1}^{N},
\end{equation}
where $q^{(i)}$ is a question and $a^{(i)} = (y_1^{(i)}, \dots, y_{T_i}^{(i)})$ is its answer token sequence. For each answer position $t$, we define the prefix context $c_t^{(i)} = [q^{(i)}; y_{<t}^{(i)}]$ and the corresponding key $k_t^{(i)} = \phi(c_t^{(i)})$, where $\phi(\cdot)$ extracts the hidden representation from a fixed teacher encoder. The datastore is constructed as
\begin{equation}
\label{eq:dstore-qa}
(\mathcal{K}, \mathcal{V})
=
\bigcup_{i=1}^{N}
\left\{(k_t^{(i)}, y_t^{(i)})\right\}_{t=1}^{T_i}.
\end{equation}
During memory training, for each SFT pair $(q,a) \in \mathcal{D}_{\mathit{sft}}$ and each answer position $t$, we form the teacher-forced decoding context $c_t = [q; y_{<t}]$. We then query the datastore with key $k_t = \phi(c_t)$, retrieve its neighbor set $\mathcal{N}(k_t)$, and construct the following non-parametric teacher distribution:
\begin{equation}
\label{eq:teacher-dist}
\begin{aligned}
p_{\mathit{teacher}}(y \mid c_t)
&\propto \sum_{(k_j, v_j) \in \mathcal{N}(k_t)}
\mathbf{1}[y = v_j]\exp \bigl(-d(k_t, k_j)/\tau\bigr),
\qquad y \in \mathcal{V},
\end{aligned}
\end{equation}
where $d(\cdot,\cdot)$ denotes a distance function and $\tau$ denotes temperature. This distribution serves as the retrieval-based teacher for memory training.

In our implementation, datastore entries follow the SFT label mask and are constructed only at answer-side positions: question tokens provide prefix context but are not stored as datastore values. We instantiate $\phi$ as the hidden state at the input to the MLP of the final decoder block, and implement nearest-neighbor search with a FAISS L2 index. Additional datastore and teacher-construction details are provided in Appendix~\ref{sec:appendix-datastore}.

\subsection{Training}
\label{sec:training}
\paragraph{Memory training}

Given the retrieval-based teacher distribution constructed above, we train the memory LM to approximate this distribution at answer-side positions. The retrieval distribution provides a richer target by assigning probability mass to multiple domain-relevant tokens \citep{xu2023nearest}, encouraging the memory to internalize domain knowledge from the datastore. For an answer token $y_t$ with prefix context $c_t = [q; y_{<t}]$, the memory model predicts $p_{\mathit{mem}}(\cdot \mid c_t)$, while the datastore provides $p_{\mathit{teacher}}(\cdot \mid c_t)$. We first minimize the Kullback-Leibler divergence between the memory's output distribution and the teacher distribution

\begin{equation}
\label{eq:mem-kl}
\mathcal{L}_{\mathit{KL}}(c_t)
=
\mathrm{KL}\!\left(
p_{\mathit{teacher}}(\cdot \mid c_t)
\;\|\;
p_{\mathit{mem}}(\cdot \mid c_t)
\right),
\end{equation}
and combine it with the cross-entropy loss
\begin{equation}
\label{eq:mem-ce}
\mathcal{L}_{\mathit{CE}}(c_t)
=
- 
\log p_{\mathit{mem}}(y_t \mid c_t).
\end{equation}
The final memory training loss is
\begin{equation}
\label{eq:mem-loss}
\mathcal{L}_{\mathit{mem}}(c_t)
=
\beta \cdot \mathcal{L}_{\mathit{KL}}(c_t)
+
(1-\beta) \cdot \mathcal{L}_{\mathit{CE}}(c_t),
\end{equation}
where $\beta \in [0,1]$ controls the balance between the retrieval-based teacher distribution and the gold SFT answer. The KL term aligns the memory LM with the datastore-derived teacher distribution, while the cross-entropy term anchors it to the gold answer token.

\paragraph{Router training}
The external memory serves as a domain expert during generation, but uniformly applying it at every decoding step can perturb the frozen backbone's general reasoning and instruction-following behavior. We therefore train a token-level router, a lightweight two-layer MLP, to predict how much the memory should contribute at each token.
 
For the same prefix context $c_t$, the backbone and memory produce distributions $p_{\mathit{base}}(\cdot \mid c_t)$ and $p_{\mathit{mem}}(\cdot \mid c_t)$. At each step, the router takes the current hidden representations from the two frozen models, together with confidence and entropy features derived from their output distributions, and predicts $\lambda_t \in [0,1]$. The fused distribution is
\begin{equation}
\label{eq:fused-dist}
\begin{aligned}
p_{\mathit{fused}}(\cdot \mid c_t)
&=
(1-\lambda_t) \cdot p_{\mathit{base}}(\cdot \mid c_t)
+
\lambda_t \cdot p_{\mathit{mem}}(\cdot \mid c_t).
\end{aligned}
\end{equation}
We train the router on mixed general and domain instruction data. The prediction term is the cross-entropy loss on the fused distribution:
\begin{equation}
\label{eq:router-ce}
\mathcal{L}_{\mathit{CE}}(c_t)
=
- 
\log p_{\mathit{fused}}(y_t \mid c_t).
\end{equation}
We additionally apply a signed linear regularizer to the memory interpolation weight:
\begin{equation}
\label{eq:router-signed}
\mathcal{R}(c_t)
=
s_t\lambda_t,
\end{equation}
where $s_t<0$ for domain examples and $s_t>0$ for general examples. The resulting per-token router objective is
\begin{equation}
\label{eq:router-loss}
\mathcal{L}_{\mathit{router}}(c_t)
=
\mathcal{L}_{\mathit{CE}}(c_t)
+
\alpha_s\mathcal{R}(c_t).
\end{equation}
Here, $\alpha_s>0$ controls the strength of the signed regularizer. Minimizing the signed term pushes $\lambda_t$ upward on domain examples and downward on general examples, while the cross-entropy term directly optimizes the final next-token distribution. Detailed router architecture and training configurations are provided in Appendix~\ref{sec:appendix-hyperparams}.

\subsection{Inference}
\label{sec:inference}
At inference time, both the frozen backbone and the trained memory LM process the same input in parallel. For the current decoding prefix $c_t = [x; y_{<t}]$, the router predicts $\lambda_t$, and the final next-token distribution is obtained by Eq.~(\ref{eq:fused-dist}). 

MemSFT avoids overwriting the backbone LLM: specialization-specific updates are confined to the external memory and router, while the backbone parameters remain frozen. The memory provides domain-specialized token distributions, and the router controls their contribution at each decoding step. The design is also modular: because memory training is decoupled from a particular target backbone, the trained memory can be reused across frozen LLMs that share a tokenizer. In practice, the backbone provides general reasoning and instruction-following capabilities, while the memory strengthens domain-specific predictions when selected by the router.

\section{Experimental Setup}
\label{sec:experimental-setup}
We evaluate whether MemSFT can improve specialized-domain performance while maintaining general performance in modern LLMs. We describe the datasets, backbones, baselines, training details, and evaluation metrics below.

\paragraph{Domain Datasets}
We evaluate MemSFT on three specialized domains. For biology, we use Biology-Instructions (BioIns), a multi-omics sequence-understanding dataset with 21 tasks \citep{he2024biology}. BioIns includes tasks over DNA, RNA, protein, and multi-sequence inputs, with evaluation metrics covering classification, regression, and sequence-function prediction. For geoscience, we use OpenSWI, a dataset for surface-wave dispersion-curve inversion \citep{liu2025openswi}, where the input consists of period, phase velocity, and group velocity, and the target physical profile contains depth, P-wave velocity, S-wave velocity, and density. For law, we use DISC-Law-SFT data from DISC-LawLLM \citep{yue2024lawllm} for specialization and evaluate legal capability on LawBench \citep{fei2024lawbench}. Due to compute constraints, we sample 500K BioIns instances, 30K OpenSWI instances, and 55K DISC-Law-SFT instances for domain specialization. 
We select these datasets to study challenging domain adaptation scenarios that require substantial specialization beyond generic instruction tuning. BioIns and OpenSWI contain inputs and outputs dominated by biological sequences, numerical arrays, and structured scientific profiles, whose formats are far from ordinary natural language and expose clear specialization gaps in the original Qwen3 backbones. LawBench complements these scientific settings with a natural-language professional domain. Detailed dataset construction and evaluation protocols are provided in Appendix~\ref{sec:appendix-datasets}.

\paragraph{Backbones and Main Baselines}
We use Qwen3 instruction-tuned models as backbones \citep{yang2025qwen3}. Across both BioIns and OpenSWI, we evaluate Qwen3-8B, Qwen3-14B, Qwen3-32B and Qwen3-235B-A22B. On LawBench, we evaluate a single Qwen3-14B configuration as an additional validation in a natural-language professional domain. Our three baselines are the original post-trained backbone without domain adaptation, full-parameter SFT \citep{wei2021finetuned,chung2024scaling}, and LoRA \citep{hu2022lora}. For LoRA, we adopt an all-linear target-module configuration following QLoRA \citep{dettmers2023qlora}, attaching adapters to all attention and MLP projections in each Transformer block; the backbone weights are not quantized. We compare all three baselines against MemSFT. For the 8B, 14B, and 32B backbones, this yields four evaluated methods in total: original backbone, SFT, LoRA, and MemSFT. The trainable baselines are trained separately for each backbone size, whereas MemSFT trains a domain-specific memory and reuses it across different backbones within the Qwen3 family, whose models share a compatible tokenizer and output vocabulary. Due to compute constraints, we do not run SFT or LoRA on Qwen3-235B-A22B, and only report the original backbone and MemSFT; the Qwen3-235B-A22B setting is therefore used to evaluate memory reuse at scale rather than to compare against trained adaptation baselines. 

\paragraph{Additional Baselines for Mitigating Forgetting}
We further include two additional baselines aimed at mitigating forgetting in the Qwen3-14B BioIns setting. First, we evaluate LoRA + MixTraining(1:1), which mixes BioIns examples with general instruction data during LoRA adaptation; this follows the MixTraining baseline used in domain-specific LLM adaptation \citep{liu2024more} and is motivated by observations that general instruction tuning can alleviate forgetting in LLM continual fine-tuning \citep{luo2025empirical}. Second, we evaluate Wise-FT-style weight-space interpolation between the original backbone and the BioIns LoRA model \citep{wortsman2022robust}.

\paragraph{Training Details}
We set the maximum sequence length to 2048 unless otherwise specified; the legal SFT and LoRA runs use 3072. For BioIns, full SFT, LoRA, and the MemSFT memory are trained for one epoch on the sampled 500K subset. For OpenSWI, the corresponding models are trained for three epochs on the sampled 30K subset. For DISC-Law, full SFT is trained for three epochs on the sampled 55K subset, while LoRA is trained for one epoch. LoRA baselines use rank 8, alpha 16, dropout 0.05, and adapters on the attention and MLP projection layers. For MemSFT, we train one domain-specific 8B memory per domain and reuse the trained memory across Qwen3 backbones when multiple backbone sizes are evaluated. The token-level router is trained with both the backbone and memory frozen, using a small mixture of domain examples and general instruction data sampled from NVIDIA's Nemotron-Post-Training-Dataset-v1 \citep{bercovich2025llama}. Additional optimization hyperparameters, router data composition, and training artifacts are provided in Appendix~\ref{sec:appendix-hyperparams}.

\paragraph{Evaluation Metrics}
For BioIns, we report the average task score across all 21 tasks \citep{he2024biology}; this aggregate combines the task-specific metrics used by BioIns, including classification, correlation, regression, and sequence-function scores. For OpenSWI, we follow the OpenCompass generation-style evaluation built on OpenSWI and report RMSE on the shallow setting \citep{liu2025openswi,cao2026opencompass}, where lower values are better. OpenSWI provides shallow and deep settings for different depth scales; in the shallow setting, the input consists of period, phase velocity, and group velocity, and the target is a near-surface S-wave velocity profile. We do not include OpenSWI-deep in the present evaluation because it requires generating much longer and more structurally complex velocity profiles, for which the current autoregressive text-generation formulation does not reliably produce well-formed outputs. For law, we report the LawBench average over the evaluated legal subtasks \citep{fei2024lawbench}.

To measure general capability retention, we use the lm-evaluation-harness framework \citep{eval-harness} and assess performance on five benchmarks: MATH-500 \citep{lightman2024let}, C-Eval \citep{huang2023c}, IFEval \citep{zhou2023instruction}, MMLU-Redux \citep{gema2025we}, and INCLUDE \citep{romanou2025include}. We follow the evaluation settings of the Qwen3 Technical Report \citep{yang2025qwen3} for consistency and comparability. MATH-500 reports math verification accuracy, IFEval reports prompt-strict accuracy, and C-Eval, MMLU-Redux, and INCLUDE report accuracy. Results obtained with stochastic decoding are averaged over five runs; detailed evaluation settings are provided in Appendix~\ref{sec:appendix-datasets}. All general-benchmark scores are shown on a 0--100 scale, where higher is better. All reported general averages and their deltas are computed from unrounded benchmark scores.

\section{Results}
\label{sec:results}

We organize the results around three levels of evidence. First, BioIns serves as the primary setting for a detailed analysis of domain specialization and general capability retention: we report the main Qwen3 scaling results, compare against forgetting-mitigation baselines, and quantify adaptation compute across backbones (Sections~\ref{sec:results-bioins} and~\ref{sec:results-flops}); we also include a non-Qwen backbone-family validation (Section~\ref{sec:results-llama2}). Second, we evaluate OpenSWI as an additional structured scientific generation setting, testing whether MemSFT also improves numerical profile generation while preserving general capabilities (Section~\ref{sec:results-openswi}). Third, we validate the same memory-router design on LawBench to examine whether the pattern extends to a natural-language professional domain (Section~\ref{sec:results-law}).

\begin{table}[H]
\centering
\footnotesize
\setlength{\tabcolsep}{3pt}
\resizebox{\linewidth}{!}{%
\begin{tabular}{llcccccl}
\toprule
\headtworow{Model} & \multicolumn{1}{c}{\headtworow{BioIns Avg.}} & \multicolumn{6}{c}{General Benchmarks} \\
\cmidrule(lr){3-8}
 & & MATH-500 & C-Eval & IFEval & MMLU-Redux & INCLUDE & \multicolumn{1}{c}{Avg.} \\
\midrule
Qwen3-8B & \rawavgcell{5.07} & \rawscorecell{95.5} & \rawscorecell{79.3} & \rawscorecell{84.5} & \rawscorecell{88.0} & \rawscorecell{55.5} & \rawavgcell{80.56} \\
+ SFT & \avgcell{37.34}{\gain{+32.26}} & \scorecell{85.2}{\drop{-10.3}} & \scorecell{77.0}{\drop{-2.3}} & \scorecell{47.3}{\drop{-37.2}} & \scorecell{74.9}{\drop{-13.1}} & \scorecell{50.8}{\drop{-4.7}} & \avgcell{67.04}{\drop{-13.52}} \\
+ LoRA & \avgcell{32.07}{\gain{+27.00}} & \scorecell{77.5}{\drop{-18.0}} & \scorecell{77.5}{\drop{-1.8}} & \scorecell{32.5}{\drop{-52.0}} & \scorecell{76.8}{\drop{-11.2}} & \scorecell{53.1}{\drop{-2.4}} & \avgcell{63.48}{\drop{-17.08}} \\
\textbf{+ MemSFT} & \avgcell{\textbf{42.84}}{\gain{\textbf{+37.77}}} & \scorecell{\textbf{96.4}}{\gain{+1.0}} & \scorecell{\textbf{79.2}}{\neutral{-0.1}} & \scorecell{\textbf{84.1}}{\neutral{-0.4}} & \scorecell{\textbf{87.8}}{\neutral{-0.2}} & \scorecell{\textbf{58.2}}{\gain{+2.7}} & \avgcell{\textbf{81.15}}{\gain{\textbf{+0.59}}} \\
\midrule
Qwen3-14B & \rawavgcell{6.64} & \rawscorecell{96.4} & \rawscorecell{83.1} & \rawscorecell{85.7} & \rawscorecell{90.0} & \rawscorecell{60.9} & \rawavgcell{83.22} \\
+ SFT & \avgcell{41.16}{\gain{+34.52}} & \scorecell{31.2}{\drop{-65.3}} & \scorecell{79.7}{\drop{-3.4}} & \scorecell{18.9}{\drop{-66.8}} & \scorecell{72.7}{\drop{-17.3}} & \scorecell{56.7}{\drop{-4.2}} & \avgcell{51.83}{\drop{-31.39}} \\
+ LoRA & \avgcell{32.07}{\gain{+25.43}} & \scorecell{96.0}{\neutral{-0.4}} & \scorecell{81.7}{\drop{-1.4}} & \scorecell{44.0}{\drop{-41.7}} & \scorecell{88.7}{\drop{-1.4}} & \scorecell{57.7}{\drop{-3.2}} & \avgcell{73.61}{\drop{-9.61}} \\
\textbf{+ MemSFT} & \avgcell{\textbf{42.92}}{\gain{\textbf{+36.28}}} & \scorecell{\textbf{96.5}}{\neutral{+0.0}} & \scorecell{\textbf{83.0}}{\neutral{-0.1}} & \scorecell{\textbf{85.4}}{\neutral{-0.3}} & \scorecell{\textbf{90.5}}{\neutral{+0.5}} & \scorecell{\textbf{62.8}}{\gain{+1.9}} & \avgcell{\textbf{83.62}}{\neutral{\textbf{+0.40}}} \\
\midrule
Qwen3-32B & \rawavgcell{6.25} & \rawscorecell{96.9} & \rawscorecell{86.7} & \rawscorecell{84.6} & \rawscorecell{91.6} & \rawscorecell{67.2} & \rawavgcell{85.39} \\
+ SFT & \avgcell{\textbf{43.75}}{\gain{\textbf{+37.50}}} & \scorecell{68.9}{\drop{-28.0}} & \scorecell{82.8}{\drop{-3.9}} & \scorecell{30.9}{\drop{-53.7}} & \scorecell{57.0}{\drop{-34.6}} & \scorecell{59.5}{\drop{-7.7}} & \avgcell{59.81}{\drop{-25.58}} \\
+ LoRA & \avgcell{41.36}{\gain{+35.11}} & \scorecell{95.8}{\drop{-1.0}} & \scorecell{85.8}{\drop{-0.9}} & \scorecell{71.9}{\drop{-12.7}} & \scorecell{90.7}{\drop{-0.9}} & \scorecell{64.0}{\drop{-3.2}} & \avgcell{81.65}{\drop{-3.74}} \\
\textbf{+ MemSFT} & \avgcell{42.82}{\gain{+36.57}} & \scorecell{\textbf{96.9}}{\neutral{+0.0}} & \scorecell{\textbf{86.5}}{\neutral{-0.2}} & \scorecell{\textbf{84.4}}{\neutral{-0.2}} & \scorecell{\textbf{91.7}}{\neutral{+0.1}} & \scorecell{\textbf{67.2}}{\neutral{+0.0}} & \avgcell{\textbf{85.33}}{\neutral{\textbf{-0.06}}} \\
\midrule
Qwen3-235B-A22B & \rawavgcell{1.14} & \rawscorecell{96.1} & \rawscorecell{89.8} & \rawscorecell{84.8} & \rawscorecell{93.4} & \rawscorecell{71.2} & \rawavgcell{87.07} \\
\textbf{+ MemSFT} & \avgcell{\textbf{42.05}}{\gain{\textbf{+40.91}}} & \scorecell{\textbf{96.6}}{\neutral{+0.5}} & \scorecell{\textbf{89.9}}{\neutral{+0.1}} & \scorecell{\textbf{84.8}}{\neutral{-0.1}} & \scorecell{\textbf{93.4}}{\neutral{+0.0}} & \scorecell{\textbf{70.9}}{\neutral{-0.3}} & \avgcell{\textbf{87.11}}{\neutral{\textbf{+0.03}}} \\
\bottomrule
\end{tabular}
}
\caption{Main results on Biology-Instructions (BioIns) and general benchmarks. Rows named by model denote the original post-trained Qwen3 backbone before domain adaptation. Colored numbers on the right denote absolute point changes relative to the corresponding model row; gray indicates negligible changes smaller than 0.5 points in magnitude. Averages and deltas are computed from unrounded scores before display rounding. MATH-500 reports math verification accuracy, and IFEval reports prompt-strict accuracy. BioIns is evaluated deterministically. All scores are reported on a 0--100 scale.}
\label{tab:bioins-main}
\end{table}

\subsection{Biology-Instructions Results}
\label{sec:results-bioins}
Table~\ref{tab:bioins-main} reports BioIns and general benchmark performance across Qwen3 backbones. All deltas are computed relative to the original backbone of the same size.

\paragraph{Domain specialization}
Original Qwen3 backbones perform poorly on BioIns despite their strong general performance, showing that the biology instruction domain requires additional specialization. MemSFT provides large and stable domain gains across all evaluated model sizes. Full SFT can also improve domain performance and slightly exceeds MemSFT on 32B, but this gain comes at a large cost to general capability.

\paragraph{General capability retention}
Full SFT substantially reduces the general average, with the largest failures on MATH-500 and IFEval, indicating that narrow domain specialization can severely damage mathematical reasoning and instruction-following behavior outside the target domain. This severe drop is partly attributable to the unusual form of BioIns: many tasks are centered on DNA, RNA, or protein sequences, whose token patterns differ sharply from natural-language instruction data. Directly optimizing the backbone on such sequence-heavy supervision can therefore shift the model away from the linguistic and instruction-following distributions acquired during post-training. LoRA can alleviate this forgetting, but it remains below MemSFT in both BioIns performance and general retention. In contrast, MemSFT achieves strong BioIns specialization with negligible degradation in general performance, avoiding the severe forgetting observed in full SFT.

\paragraph{Scaling across Qwen3 backbones}
MemSFT decouples memory training from the backbone, allowing the same 8B memory to be reused across frozen Qwen3 backbones up to Qwen3-235B-A22B. As the backbone scales, MemSFT maintains strong domain gains and consistent general retention, supporting its scalability as a plug-in memory module for modern LLMs.

\Needspace{15\baselineskip}
\begin{wrapfigure}{r}{0.49\linewidth}
\centering
\includegraphics[width=\linewidth]{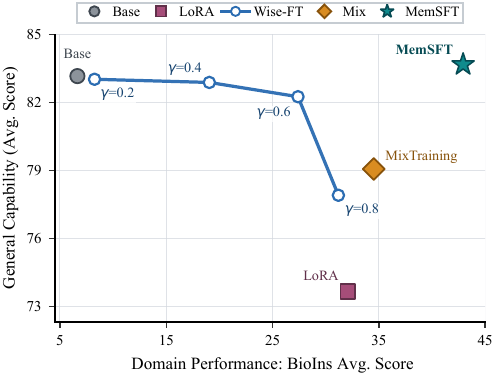}
\caption{Forgetting-mitigation baselines on BioIns with Qwen3-14B.}
\label{fig:bioins-antiforgetting}
\end{wrapfigure}
\paragraph{Baselines for mitigating forgetting}
We further test whether standard strategies for mitigating forgetting can match MemSFT in both domain performance and general capability retention on Qwen3-14B. Figure~\ref{fig:bioins-antiforgetting} compares two targeted baselines: LoRA + MixTraining(1:1), which mixes the 500K BioIns subset with 500K general instruction examples, and Wise-FT-style weight-space interpolation, which interpolates between the original backbone and the BioIns LoRA model by scaling the LoRA update. Wise-FT shows a clear interpolation curve: intermediate interpolation weights offer a reasonable balance between BioIns improvement and general retention, but no point on the curve approaches MemSFT. MixTraining also mitigates LoRA forgetting, but it does not eliminate the trade-off and remains below MemSFT despite using additional general data and more optimization steps. Together, these baselines show that standard replay and weight interpolation are insufficient to match MemSFT's combination of strong domain performance and general capability retention. Full benchmark results and implementation details for these baselines are provided in Appendix~\ref{sec:appendix-antiforgetting}.

\subsection{Adaptation Compute Across Backbones}
\label{sec:results-flops}

We estimate the analytical model FLOPs required to specialize Qwen3 backbones on BioIns before deployment. Following standard Transformer FLOPs accounting, we use $6P_{\mathrm{act}}T$ for full SFT and an optimistic lower-bound estimate of $4P_{\mathrm{act}}T$ for LoRA, where $P_{\mathrm{act}}$ is the number of parameters activated per token and $T$ is the number of specialization tokens. For the dense Qwen3 backbones, we use their nominal parameter counts; for Qwen3-235B-A22B, we use its 22B activated parameters rather than its 235B total parameters. For MemSFT, we include one Qwen3-8B forward pass to construct the datastore, one shared 8B memory-training run, and backbone-specific router training with both the backbone and memory frozen. A detailed derivation is provided in Appendix~\ref{sec:appendix-flops}. Reusing the shared domain memory across more backbones further reduces MemSFT's relative adaptation cost; the 0.96-EFLOP router cost represents only the marginal cost of attaching an already trained memory, not the full cost of standalone 235B specialization.

\begin{table}[H]
\centering
\footnotesize
\begin{minipage}[t]{0.56\linewidth}
\centering
\textbf{(a) Adaptation cost comparison}\par
\vspace{2pt}
\setlength{\tabcolsep}{3pt}
\begin{tabular*}{\linewidth}{@{\extracolsep{\fill}}lcccc@{}}
\toprule
& \multicolumn{2}{c}{235B standalone} & \multicolumn{2}{c}{Four backbones} \\
\cmidrule(lr){2-3}\cmidrule(l){4-5}
Method & EFLOPs & \cellcolor{hdr}\textbf{Rel. $\downarrow$} & EFLOPs & \cellcolor{hdr}\textbf{Rel. $\downarrow$} \\
\midrule
SFT & 11.88 & \cellcolor{hdr}1.00x & 41.05 & \cellcolor{hdr}1.00x \\
LoRA & 7.92 & \cellcolor{hdr}0.67x & 27.37 & \cellcolor{hdr}0.67x \\
\textbf{MemSFT} & 6.73 & \cellcolor{tablegreen}\textbf{0.57x} & 9.23 & \cellcolor{tablegreen}\textbf{0.22x} \\
\bottomrule
\end{tabular*}
\end{minipage}
\hfill
\begin{minipage}[t]{0.41\linewidth}
\centering
\textbf{(b) MemSFT cost breakdown}\par
\vspace{2pt}
\setlength{\tabcolsep}{3pt}
\begin{tabular*}{\linewidth}{@{\extracolsep{\fill}}lcc@{}}
\toprule
Stage & 235B & All four \\
\midrule
Datastore & 1.44 & 1.44 \\
8B memory training & 4.32 & 4.32 \\
Routers & 0.96 & 3.47 \\
\midrule
\textbf{Total} & \textbf{6.73} & \textbf{9.23} \\
\bottomrule
\end{tabular*}
\end{minipage}
\caption{Estimated model FLOPs for Biology-Instructions adaptation. The standalone Qwen3-235B-A22B setting includes all MemSFT stages, whereas the four-backbone setting shares datastore construction and memory training across separate backbone-specific routers. Relative costs are normalized by SFT within each scope; Qwen3-235B-A22B uses 22B activated parameters per token. Totals are computed from unrounded estimates.}
\label{tab:adaptation-flops}
\end{table}

\Needspace{8\baselineskip}
\subsection{Additional Backbone-Family Validation}
\label{sec:results-llama2}

To test whether MemSFT is specific to the Qwen3 family, we further evaluate it on LLaMA2 under the BioIns setting. Table~\ref{tab:llama2-validation} reports a validation with a LLaMA2-13B backbone and a 7B memory. MemSFT substantially improves BioIns performance over the original backbone while keeping the general benchmark average nearly unchanged. In contrast, full SFT achieves strong domain performance but substantially degrades general capabilities.

\begin{table}[H]
\centering
\footnotesize
\setlength{\tabcolsep}{3pt}
\resizebox{\linewidth}{!}{%
\begin{tabular}{lccccccc}
\toprule
\headtworow{Model} & \headtworow{BioIns Avg.} & \multicolumn{6}{c}{General Benchmarks} \\
\cmidrule(lr){3-8}
 & & MATH-500 & C-Eval & IFEval & MMLU-Redux & INCLUDE & \multicolumn{1}{c}{Avg.} \\
\midrule
LLaMA2-13B & \rawavgcell{-18.28} & \rawscorecell{7.1} & \rawscorecell{38.5} & \rawscorecell{32.9} & \rawscorecell{53.2} & \rawscorecell{37.0} & \rawavgcell{33.74} \\
+ SFT & \avgcell{32.45}{\gain{+50.72}} & \scorecell{1.6}{\drop{-5.5}} & \scorecell{26.2}{\drop{-12.3}} & \scorecell{13.9}{\drop{-19.0}} & \scorecell{43.9}{\drop{-9.3}} & \scorecell{26.9}{\drop{-10.2}} & \avgcell{22.47}{\drop{-11.27}} \\
+ LoRA & \avgcell{10.35}{\gain{+28.63}} & \scorecell{6.2}{\drop{-1.0}} & \scorecell{25.3}{\drop{-13.2}} & \scorecell{28.8}{\drop{-4.1}} & \scorecell{50.4}{\drop{-2.8}} & \scorecell{34.4}{\drop{-2.6}} & \avgcell{29.01}{\drop{-4.73}} \\
\textbf{+ MemSFT} & \avgcell{\textbf{32.95}}{\gain{\textbf{+51.23}}} & \scorecell{\textbf{7.4}}{\neutral{\textbf{+0.2}}} & \scorecell{\textbf{38.3}}{\neutral{\textbf{-0.1}}} & \scorecell{\textbf{33.1}}{\neutral{\textbf{+0.2}}} & \scorecell{\textbf{53.2}}{\neutral{\textbf{0.0}}} & \scorecell{\textbf{37.1}}{\neutral{\textbf{+0.0}}} & \avgcell{\textbf{33.80}}{\neutral{\textbf{+0.06}}} \\
\bottomrule
\end{tabular}
}
\caption{Additional backbone-family validation on Biology-Instructions (BioIns) with LLaMA2-13B. MemSFT uses a 7B memory and preserves the original backbone's general performance while improving domain performance. BioIns Avg. can be negative because the aggregate includes correlation-style metrics such as MCC and $R^2$.}
\label{tab:llama2-validation}
\end{table}

\subsection{OpenSWI Results}
\label{sec:results-openswi}
Table~\ref{tab:openswi-main} reports OpenSWI results under the shallow generation-style evaluation protocol, together with general benchmark performance across Qwen3 backbones. For general benchmarks, deltas are computed relative to the original backbone of the same size. For RMSE, negative deltas indicate improvement because lower values are better.

\begin{table}[H]
\centering
\footnotesize
\renewcommand{\arraystretch}{0.9}
\setlength{\tabcolsep}{3pt}
\resizebox{\textwidth}{!}{%
\begin{tabular}{lccccccc}
\toprule
\headtworow{Model} & \multicolumn{1}{c}{OpenSWI} & \multicolumn{6}{c}{General Benchmarks} \\
\cmidrule(lr){2-2}\cmidrule(lr){3-8}
 & RMSE $\downarrow$ & MATH-500 & C-Eval & IFEval & MMLU-Redux & INCLUDE & \multicolumn{1}{c}{Avg.} \\
\midrule
Qwen3-8B & \rawavgcell{103.07} & \rawscorecell{95.5} & \rawscorecell{79.3} & \rawscorecell{84.5} & \rawscorecell{88.0} & \rawscorecell{55.5} & \rawavgcell{80.56} \\
+ SFT & \avgcell{0.51}{\gain{-102.56}} & \scorecell{51.7}{\drop{-43.8}} & \scorecell{78.5}{\drop{-0.8}} & \scorecell{32.8}{\drop{-51.7}} & \scorecell{64.6}{\drop{-23.4}} & \scorecell{55.1}{\neutral{-0.4}} & \avgcell{56.54}{\drop{-24.02}} \\
+ LoRA & \avgcell{0.52}{\gain{-102.55}} & \scorecell{71.8}{\drop{-23.7}} & \scorecell{79.0}{\neutral{-0.3}} & \scorecell{77.1}{\drop{-7.4}} & \scorecell{55.6}{\drop{-32.4}} & \scorecell{55.2}{\neutral{-0.3}} & \avgcell{67.74}{\drop{-12.82}} \\
\textbf{+ MemSFT} & \avgcell{\textbf{0.47}}{\gain{\textbf{-102.60}}} & \scorecell{\textbf{95.8}}{\neutral{\textbf{+0.3}}} & \scorecell{\textbf{79.2}}{\neutral{\textbf{-0.1}}} & \scorecell{\textbf{84.7}}{\neutral{\textbf{+0.1}}} & \scorecell{\textbf{87.9}}{\neutral{\textbf{-0.1}}} & \scorecell{\textbf{58.2}}{\gain{\textbf{+2.7}}} & \avgcell{\textbf{81.13}}{\gain{\textbf{+0.57}}} \\
\midrule
Qwen3-14B & \rawavgcell{4.39} & \rawscorecell{96.4} & \rawscorecell{83.1} & \rawscorecell{85.7} & \rawscorecell{90.0} & \rawscorecell{60.9} & \rawavgcell{83.22} \\
+ SFT & \avgcell{0.62}{\gain{-3.77}} & \scorecell{57.7}{\drop{-38.7}} & \scorecell{82.4}{\drop{-0.7}} & \scorecell{44.5}{\drop{-41.1}} & \scorecell{45.8}{\drop{-44.2}} & \scorecell{60.3}{\drop{-0.6}} & \avgcell{58.15}{\drop{-25.07}} \\
+ LoRA & \avgcell{0.80}{\gain{-3.59}} & \scorecell{65.4}{\drop{-31.0}} & \scorecell{82.7}{\neutral{-0.4}} & \scorecell{81.7}{\drop{-4.0}} & \scorecell{82.9}{\drop{-7.2}} & \scorecell{60.5}{\neutral{-0.4}} & \avgcell{74.63}{\drop{-8.59}} \\
\textbf{+ MemSFT} & \avgcell{\textbf{0.47}}{\gain{\textbf{-3.92}}} & \scorecell{\textbf{96.7}}{\neutral{\textbf{+0.3}}} & \scorecell{\textbf{83.4}}{\neutral{\textbf{+0.3}}} & \scorecell{\textbf{86.1}}{\neutral{\textbf{+0.5}}} & \scorecell{\textbf{90.0}}{\neutral{\textbf{+0.0}}} & \scorecell{\textbf{62.8}}{\gain{\textbf{+1.9}}} & \avgcell{\textbf{83.80}}{\gain{\textbf{+0.58}}} \\
\midrule
Qwen3-32B & \rawavgcell{1.40} & \rawscorecell{96.9} & \rawscorecell{86.7} & \rawscorecell{84.6} & \rawscorecell{91.6} & \rawscorecell{67.2} & \rawavgcell{85.39} \\
+ SFT & \avgcell{0.49}{\gain{-0.91}} & \scorecell{65.2}{\drop{-31.7}} & \scorecell{84.8}{\drop{-1.9}} & \scorecell{26.3}{\drop{-58.3}} & \scorecell{77.3}{\drop{-14.3}} & \scorecell{65.1}{\drop{-2.1}} & \avgcell{63.73}{\drop{-21.66}} \\
+ LoRA & \avgcell{0.50}{\gain{-0.90}} & \scorecell{73.2}{\drop{-23.7}} & \scorecell{86.9}{\neutral{+0.2}} & \scorecell{80.3}{\drop{-4.3}} & \scorecell{85.7}{\drop{-5.9}} & \scorecell{66.0}{\drop{-1.2}} & \avgcell{78.42}{\drop{-6.97}} \\
\textbf{+ MemSFT} & \avgcell{\textbf{0.47}}{\gain{\textbf{-0.93}}} & \scorecell{\textbf{97.0}}{\neutral{\textbf{+0.1}}} & \scorecell{\textbf{86.6}}{\neutral{\textbf{-0.1}}} & \scorecell{\textbf{84.6}}{\neutral{\textbf{+0.0}}} & \scorecell{\textbf{91.6}}{\neutral{\textbf{+0.0}}} & \scorecell{\textbf{67.2}}{\neutral{\textbf{+0.0}}} & \avgcell{\textbf{85.41}}{\neutral{\textbf{+0.02}}} \\
\midrule
Qwen3-235B-A22B & \rawavgcell{0.96} & \rawscorecell{96.1} & \rawscorecell{89.8} & \rawscorecell{84.8} & \rawscorecell{93.4} & \rawscorecell{71.2} & \rawavgcell{87.07} \\
\textbf{+ MemSFT} & \avgcell{\textbf{0.47}}{\gain{\textbf{-0.49}}} & \scorecell{\textbf{95.7}}{\neutral{\textbf{-0.4}}} & \scorecell{\textbf{89.6}}{\neutral{\textbf{-0.2}}} & \scorecell{\textbf{84.2}}{\drop{\textbf{-0.6}}} & \scorecell{\textbf{93.5}}{\neutral{\textbf{+0.1}}} & \scorecell{\textbf{73.1}}{\gain{\textbf{+1.9}}} & \avgcell{\textbf{87.22}}{\neutral{\textbf{+0.15}}} \\
\bottomrule
\end{tabular}
}
\caption{Results on OpenSWI and general benchmarks. OpenSWI is evaluated under the shallow generation-style protocol, using RMSE over generated shallow S-wave velocity profiles. Rows named by model denote the original post-trained Qwen3 backbone before domain adaptation. Colored numbers on the right denote absolute changes relative to the corresponding model row. For RMSE, negative deltas indicate improvement because lower values are better; for general-benchmark deltas, gray indicates negligible changes smaller than 0.5 points in magnitude.}
\label{tab:openswi-main}
\end{table}

\paragraph{OpenSWI task performance}
Original Qwen3 backbones struggle on OpenSWI, with especially large error at 8B and still substantial error at 14B. Full SFT and LoRA both reduce the prediction error, but MemSFT achieves the lowest RMSE on 8B, 14B, and 32B. This indicates that the external memory substantially improves structured geological prediction without rewriting the backbone.

\paragraph{General capability retention}
The OpenSWI results reproduce the domain gains and general capability retention observed on BioIns. Full SFT sharply reduces general performance, with especially severe drops on IFEval and MMLU-Redux. LoRA alleviates part of this degradation, but it still incurs noticeable losses on MATH-500 and MMLU-Redux. In contrast, MemSFT keeps the general average nearly unchanged across all evaluated backbone sizes while still delivering strong OpenSWI gains.

\subsection{Legal-Domain Validation}
\label{sec:results-law}

We evaluate MemSFT on legal-domain specialization using a Qwen3-14B backbone with an 8B legal memory. As shown in Table~\ref{tab:lawbench}, SFT and LoRA improve LawBench performance but substantially degrade general capabilities, especially on MATH-500 and IFEval. MemSFT reaches a LawBench score comparable to the strongest trainable baseline while preserving the general benchmark average, suggesting that the memory-router design also transfers to professional textual domains.

\begin{table}[H]
\centering
\footnotesize
\setlength{\tabcolsep}{3pt}
\resizebox{\linewidth}{!}{%
\begin{tabular}{llcccccl}
\toprule
\headtworow{Model} & \multicolumn{1}{c}{\headtworow{LawBench Avg.}} & \multicolumn{6}{c}{General Benchmarks} \\
\cmidrule(lr){3-8}
 & & MATH-500 & C-Eval & IFEval & MMLU-Redux & INCLUDE & \multicolumn{1}{c}{Avg.} \\
\midrule
Qwen3-14B & \rawavgcell{49.83} & \rawscorecell{96.4} & \rawscorecell{83.1} & \rawscorecell{85.7} & \rawscorecell{90.0} & \rawscorecell{60.9} & \rawavgcell{83.22} \\
+ SFT & \avgcell{55.03}{\gain{+5.20}} & \scorecell{67.5}{\drop{-29.0}} & \scorecell{81.6}{\drop{-1.5}} & \scorecell{70.1}{\drop{-15.6}} & \scorecell{83.0}{\drop{-7.0}} & \scorecell{62.6}{\gain{+1.7}} & \avgcell{72.96}{\drop{-10.26}} \\
+ LoRA & \avgcell{56.45}{\gain{+6.62}} & \scorecell{51.8}{\drop{-44.6}} & \scorecell{83.4}{\neutral{+0.3}} & \scorecell{70.5}{\drop{-15.2}} & \scorecell{81.5}{\drop{-8.5}} & \scorecell{61.4}{\gain{+0.5}} & \avgcell{69.71}{\drop{-13.51}} \\
\textbf{+ MemSFT} & \avgcell{\textbf{56.47}}{\gain{\textbf{+6.64}}} & \scorecell{\textbf{96.7}}{\neutral{\textbf{+0.3}}} & \scorecell{\textbf{83.0}}{\neutral{\textbf{-0.1}}} & \scorecell{\textbf{86.7}}{\gain{\textbf{+1.0}}} & \scorecell{\textbf{89.8}}{\neutral{\textbf{-0.2}}} & \scorecell{\textbf{62.8}}{\gain{\textbf{+1.9}}} & \avgcell{\textbf{83.79}}{\gain{\textbf{+0.57}}} \\
\bottomrule
\end{tabular}
}
\caption{Additional validation on LawBench. LawBench Avg. is computed over the 19 evaluated LawBench subtasks under the adopted evaluation protocol; full per-task results are provided in Appendix~\ref{sec:appendix-per-task-domain}.}
\label{tab:lawbench}
\end{table}

\section{Analysis}
\label{sec:analysis}

\subsection{Learned Token-Level Routing vs. Fixed Interpolation}
\label{sec:analysis-router}
\begin{figure}[!t]
\centering
\begin{subfigure}[t]{0.485\linewidth}
    \centering
    \includegraphics[width=\linewidth]{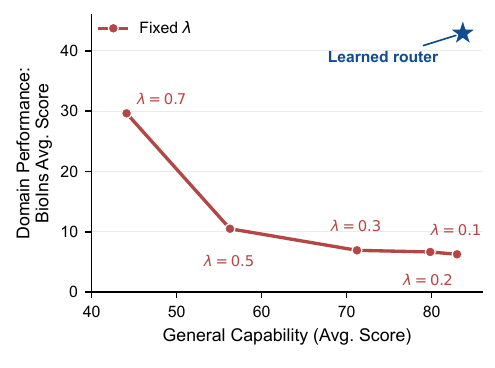}
    \captionsetup{justification=centering,singlelinecheck=false}
    \caption{Learned routing vs. fixed interpolation.}
    \label{fig:router-fixed-lambda}
\end{subfigure}
\hfill
\begin{subfigure}[t]{0.485\linewidth}
    \centering
    \includegraphics[width=\linewidth]{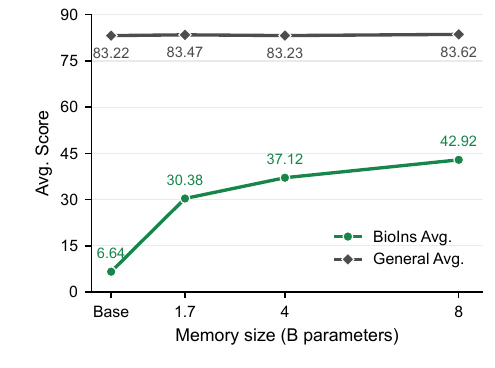}
    \captionsetup{justification=centering,singlelinecheck=false}
    \caption{Effect of memory scale.}
    \label{fig:memory-scale}
\end{subfigure}
\caption{Design analyses on Biology-Instructions (BioIns) with Qwen3-14B. (a) Fixed interpolation traces a trade-off between domain performance and general capability retention, whereas learned token-level routing achieves strong domain performance while preserving general capability. (b) Increasing memory size improves BioIns performance while the general average remains nearly unchanged. ``Base'' denotes the frozen backbone without an external memory. All memory sizes use the same Biology-Instructions training data, Qwen3-8B retrieval teacher, frozen Qwen3-14B backbone, and memory-router pipeline, with size-specific optimization configurations.}
\label{fig:router-memory-analysis}
\end{figure}

We compare the learned token-level router against fixed interpolation coefficients under the same Qwen3-14B backbone and Qwen3-8B memory. As shown in Figure~\ref{fig:router-fixed-lambda}, fixed interpolation creates a sharp trade-off between domain performance and general capability retention. With a small interpolation coefficient, the model remains close to the backbone but barely uses the memory: $\lambda=0.1$ keeps the general average at 83.21 but reaches only 6.28 on BioIns. Increasing the coefficient improves domain performance but progressively damages general capability; at $\lambda=0.7$, BioIns rises to 29.65 while the general average falls to 44.34.

\Needspace{6\baselineskip}
The learned router avoids this global trade-off by predicting a token-wise interpolation coefficient $\lambda_t$ rather than applying a constant coefficient across all decoding steps. It reaches 42.92 on BioIns while maintaining an 83.62 general average, placing it well beyond the observed fixed-interpolation trade-off. These results show that learned token-wise fusion is substantially more effective than a single global interpolation coefficient in this setting. Rather than assigning the same memory strength throughout generation, the router can vary the memory contribution across decoding steps; Section~\ref{sec:analysis-router-usage} provides a qualitative case study of this behavior.

\subsection{Effect of Memory Scale on Specialization}
\label{sec:analysis-memory-scale}
Under the BioIns setting, we study how the size of the external memory affects specialization with the same frozen Qwen3-14B backbone. Figure~\ref{fig:memory-scale} compares 1.7B, 4B, and 8B memory models attached to Qwen3-14B. Larger memories provide stronger BioIns adaptation: the 1.7B memory improves BioIns from 6.64 to 30.38, the 4B memory further increases it to 37.12, and the 8B memory reaches 42.92. At the same time, all memory sizes preserve general performance, with general averages ranging from 83.23 to 83.62.

These results suggest that increasing memory size improves the domain-specialized capacity available to MemSFT, while the frozen backbone and router preserve general capability. The 8B memory achieves the strongest BioIns performance, while the 1.7B and 4B memories still improve the backbone by 23.74 and 30.48 points, respectively, with comparable general-capability retention. This result shows a consistent capacity--performance trend under the shared training pipeline without changing the frozen backbone.

\subsection{Case Study: The Router Selectively Uses Memory}
\label{sec:analysis-router-usage}
\begin{figure}[H]
    \centering
    \begin{minipage}{0.965\textwidth}
    \begin{tcolorbox}[
        colback=orange!3!white,
        colframe=orange!65!white,
        boxrule=0.8pt,
        arc=10pt,
        outer arc=10pt,
        left=8pt,
        right=8pt,
        top=6pt,
        bottom=7pt
    ]
        \small
        \noindent Answer the question:\hfill
        \colorbox{orange!18}{\strut\textbf{OpenSWI Case}}

        \vspace{0.35em}
        \noindent\textbf{Question:} Given surface-wave dispersion arrays for period $[0.2000,\ldots]$, phase velocity $[\ldots]$, and group velocity $[\ldots]$, invert for the S-wave velocity at 70 depth points spanning 0.0000--2.7600 km (0.0400 km per layer). Provide the predicted profile as a Python list of 70 floats.

        \vspace{0.2em}
        \noindent\tikz{\draw[gray!65,dashed,line width=0.5pt] (0,0)--(\linewidth,0);}

        \vspace{0.35em}
        \noindent\raggedright\textbf{Output:} I'll generate a Vs sequence that starts at 0.3484 and increases linearly with depth. The depth layers are 0.0400 km, so the Vs values would be: Vs result sequence = $[0.3484,\allowbreak 0.3484,\allowbreak \ldots,\allowbreak 1.2883,\allowbreak 1.2883]$.
    \end{tcolorbox}
    \end{minipage}

    \vspace{0.7em}
    \begin{minipage}[c]{0.600\textwidth}
        \centering
        \includegraphics[width=\linewidth]{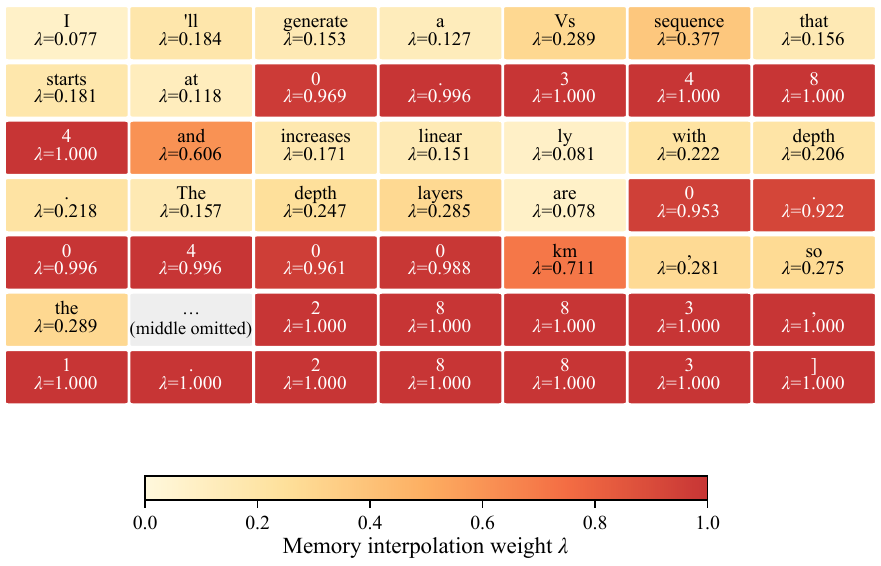}
        \vspace{-0.4em}

        \textbf{(a)} Exact coefficients for the displayed tokenizer tokens.
    \end{minipage}\hfill
    \begin{minipage}[c]{0.370\textwidth}
        \centering
        \includegraphics[width=\linewidth]{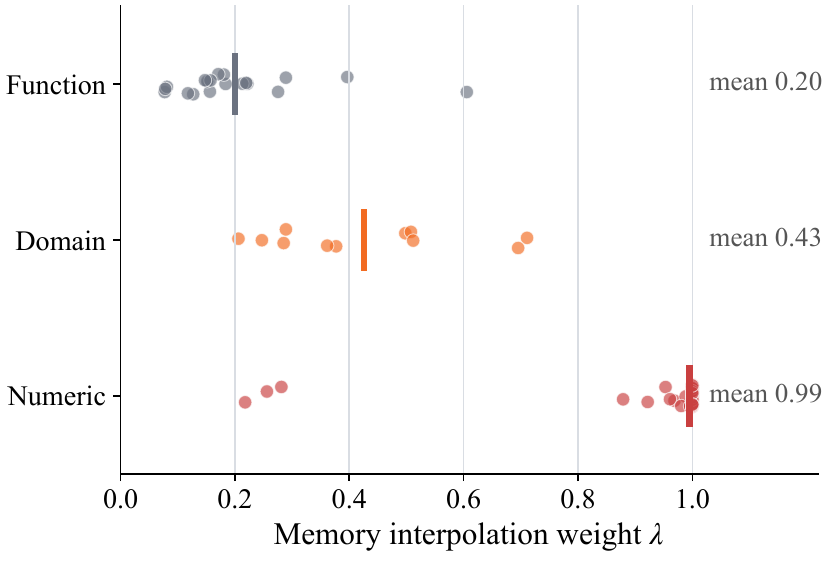}
        \vspace{-0.4em}

        \textbf{(b)} Lambda distribution.
    \end{minipage}
    \caption{Token-level router behavior on a representative OpenSWI prompt. The gray cell in (a) marks an omitted middle span, and every retained cell reports the exact coefficient of one tokenizer token. Panel (b) groups all tokens in the complete answer passage by surface role; vertical bars denote category means.}
    \label{fig:router-openswi-case}
\end{figure}

Fixed-$\lambda$ results establish that memory must be invoked selectively, but they do not reveal how that selectivity is expressed within a response. We therefore inspect a representative OpenSWI generation that combines ordinary explanatory language with a structured S-wave velocity profile. Figure~\ref{fig:router-openswi-case} presents the prompt context, exact coefficients for an abbreviated token sequence, and the distribution over the complete answer passage.

Routing changes sharply with token role within the same answer. Function tokens remain mostly backbone-dominant (mean 0.20), domain-bearing terms receive more memory weight (mean 0.43), and numerical profile tokens are routed almost entirely through memory (mean 0.99). This within-response switch illustrates the advantage over fixed interpolation in Figure~\ref{fig:router-fixed-lambda}: the model can preserve backbone reasoning while exposing memory only to spans that benefit from specialization.

Appendix~\ref{sec:appendix-router-cases} broadens this analysis with one BioIns case, two MATH cases, and two MMLU cases. The comparison separates domain relevance from surface form: the BioIns label is strongly memory-routed, ordinary mathematical numbers receive only modest weights, and the MMLU answers remain almost entirely backbone-routed. These cases illustrate that the router responds to a token's role in the task rather than to a fixed lexical category.

\Needspace{16\baselineskip}
\subsection{Comparison with Direct Retrieval}
\label{sec:analysis-rag}

\begin{wrapfigure}[14]{r}{0.48\linewidth}
\vspace{-0.6\baselineskip}
\centering
\includegraphics[width=\linewidth]{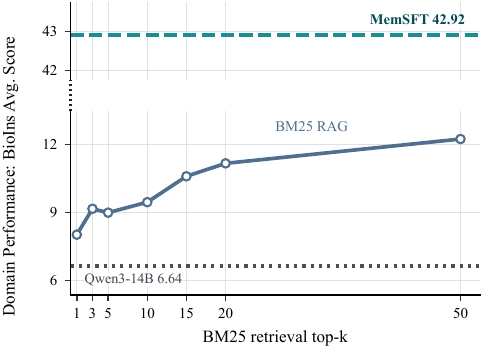}
\caption{Direct BM25 RAG on BioIns with Qwen3-14B.}
\label{fig:bioins-rag-topk}
\end{wrapfigure}
We examine whether MemSFT's domain gains can be reproduced by directly retrieving BioIns training examples at inference time. We construct a BM25 RAG baseline for Qwen3-14B using the same 500K-example BioIns subset as the retrieval corpus \citep{robertson2009probabilistic,lewis2020retrieval}. As shown in Figure~\ref{fig:bioins-rag-topk}, increasing top-$k$ improves the BioIns average from 6.64 for the original backbone to 12.23 at top-50, but this remains far below MemSFT's 42.92. Direct retrieval can expose the backbone to relevant examples, but it does not provide domain specialization comparable to the learned parametric memory. Full top-$k$ and per-task results are provided in Appendix~\ref{sec:appendix-rag}.

\FloatBarrier
\section{Related Work}
\label{sec:related}

\subsection{Domain Specialization and Forgetting}
Domain specialization has emerged as a broad direction for customizing LLMs to domain data, knowledge, objectives, and constraints \citep{ling2025domain}. A common practical route is to further optimize LLMs on target-domain supervision, such as domain-specific instruction tuning or task-oriented fine-tuning \citep{he2024biology,yu2024llasmol,yue2024lawllm}. Such full-parameter SFT is effective for specialization, but the parameter updates that make it effective can also erode the general capabilities acquired during pretraining and post-training, leading to catastrophic forgetting \citep{mccloskey1989catastrophic,li2024revisiting,luo2025empirical}. Parameter-efficient adaptation methods and strategies for mitigating forgetting, including LoRA and adapters \citep{hu2022lora,houlsby2019parameter}, mixed-domain training or replay \citep{ding2025improved,huang2024mitigating}, and weight-space interpolation such as Wise-FT \citep{wortsman2022robust}, can mitigate forgetting by reducing update size, regularizing with general data, or interpolating toward the original model. However, they remain subject to the trade-off between domain specialization and general capability retention \citep{kalajdzievski2024scaling,biderman2024lora} and typically require a separate adaptation run for each target model. In contrast, MemSFT moves beyond this trade-off by keeping the backbone frozen and externalizing domain knowledge into a reusable memory that can be transferred across compatible backbones, enabling strong domain gains with negligible degradation in general performance.

\subsection{Memory for Language Models}
\paragraph{Short-term memory.}
Short-term memory retains representations derived from the input sequence, thereby extending the context available to the model. This category includes long-context architectures and context-dependent memory systems. Transformer-XL reuses hidden states across segments \citep{dai2019transformer}. Compressive Transformer extends this mechanism by compressing past activations for longer-range access \citep{rae2019compressive}. Memorizing Transformers and LongMem retrieve representations of past inputs from external memory banks \citep{wu2022memorizing,wang2023augmenting}. MSA sparsely attends to compressed key--value states, scaling end-to-end access to extremely long contexts \citep{chen2026msa}. These approaches are orthogonal to MemSFT and can complement it when processing long documents or interactions.

\paragraph{Long-term memory.}
Long-term memory preserves information beyond the current context through external datastores or learned parametric states. It takes two main forms: non-parametric and parametric memory. \textit{Non-parametric memory} retrieves information from external datastores. Retrieval-augmented language models retrieve textual evidence from external corpora at inference time \citep{guu2020retrieval,lewis2020retrieval,borgeaud2022improving}. Memory$^3$ stores sparse explicit memories accessed through attention layers \citep{yang2024text}. $k$NN-LM interpolates token-level neighbor distributions from a datastore with the base LM \citep{khandelwal2019generalization}. MemSFT uses such a non-parametric datastore only to provide soft supervision during training, not at inference. \textit{Parametric memory} stores information in fixed-capacity learned modules. Earlier modular approaches, including K-Adapter and Mixture-of-Domain-Adapters, keep the pretrained backbone frozen and use auxiliary adapters to encode factual knowledge or domain expertise \citep{wang2021k,diao2023mixture}. Although they modularize knowledge acquisition, these adapters remain coupled to a particular backbone and primarily target pretrained language models rather than domain specialization of post-trained LLMs. Memory Layers at Scale uses trainable key--value lookup layers \citep{berges2024memory}. Engram uses deterministic $N$-gram keys to access learned memory embeddings \citep{cheng2026conditional}. Titans introduces a neural long-term memory that is updated at test time \citep{behrouz2026titans}. Nested Learning generalizes this idea to a continuum of memories with different update frequencies \citep{behrouz2025nested}. Memory Decoder and MLP Memory train reusable parametric memory modules to approximate the output distributions of non-parametric retrievers, then combine them with compatible language models using a globally tuned interpolation weight \citep{cao2026memory,wei2025mlp}. MemSFT extends this parametric memory paradigm to modern post-trained LLMs, providing domain expertise through a plug-and-play memory model that remains fixed at inference.

\section{Conclusion}
\label{sec:conclusion}

We introduced MemSFT, a modular approach that mitigates the alignment tax by decoupling domain specialization from backbone parameter updates through a plug-and-play parametric memory. Across biology, geoscience, and law, MemSFT consistently improves domain performance with negligible degradation in general performance, while allowing the same domain memory to be reused across Qwen3 backbones ranging from Qwen3-8B to Qwen3-235B-A22B. These results show that modern LLMs can be equipped with specialized capabilities without repeated backbone fine-tuning. MemSFT therefore provides a modular path for adapting models to evolving scientific and professional domains by training new domain-specific memories without frequent backbone retraining.

\section{Limitations}
\label{sec:limitations}

While MemSFT demonstrates strong domain gains with negligible degradation in general capabilities across the evaluated domains, the current study has several limitations. MemSFT currently reuses a domain memory across backbones that share a compatible tokenizer and output vocabulary, as in our Qwen3 experiments. Directly transferring the same memory across unrelated model families would require additional vocabulary alignment; cross-vocabulary adaptation through brief continued training is a promising direction but is left for future work. In addition, this paper focuses on supervised memory construction and does not explore reinforcement learning as an additional training stage. Since the memory is a full decoder model rather than a static retrieval module, future work could apply reinforcement learning to further enhance its domain-specific capability and align it better with the frozen backbone.

\section*{Acknowledgments}

This work is sponsored by the National Natural Science Foundation of China (NSFC) grant (No. 62576211) and the National Key Research and Development Program of China (No. 2023ZD0121402). It is also the result of a collaborative project on novel language model architectures between Shanghai Jiao Tong University (SJTU) and the Shanghai Artificial Intelligence Laboratory. The computational resources required for training and evaluating the models were provided by the Shanghai AI Lab. This work is also supported by the Specialized Program on Fundamental Research from Science and Technology Commission of Shanghai Municipality (No. 2025SHZDZX025G09).

\bibliography{custom}

\appendix

\section{Datasets and Evaluation Protocols}
\label{sec:appendix-datasets}

Table~\ref{tab:appendix-dataset-overview} summarizes the role and adopted scope of each domain dataset. The same sampled domain corpus is used for full SFT, LoRA, and memory training within each domain unless otherwise noted, ensuring that adaptation methods differ in training mechanism rather than access to domain supervision. For BioIns and OpenSWI, training data are sampled exclusively from the official training splits, while evaluation uses the corresponding official evaluation splits. These splits are disjoint, and no evaluation examples are used for SFT, LoRA, memory training, datastore construction, or router training.

\begin{table}[H]
\centering
\footnotesize
\setlength{\tabcolsep}{4pt}
\begin{tabularx}{\linewidth}{@{}>{\raggedright\arraybackslash}p{0.20\linewidth} l l >{\raggedright\arraybackslash}X l@{}}
\toprule
Dataset & Role & Adopted scope & Input / output form & Main metric \\
\midrule
Biology-Instructions & Train + eval. & 500K train; 21 tasks & Biological sequences and structured task answers & Task average \\
OpenSWI & Train + eval. & 30K train; shallow eval. & Dispersion curves to shallow $V_s$ profiles & RMSE $\downarrow$ \\
DISC-Law-SFT & Train & 55,295 examples & Legal instructions and responses & -- \\
LawBench & Eval. & 19 evaluated tasks & Legal knowledge and reasoning tasks & Task average \\
Nemotron-Post-Training-Dataset-v1 & Router training & 15.3K examples & General instruction data & -- \\
\bottomrule
\end{tabularx}
\caption{Datasets used for domain specialization, router training, and evaluation. Counts report the subsets adopted in this work rather than the full size of each released dataset.}
\label{tab:appendix-dataset-overview}
\end{table}

\subsection{Domain Specialization and Evaluation Data}

\paragraph{Biology-Instructions.}
Biology-Instructions contains 21 tasks spanning DNA, RNA, protein, and multi-sequence understanding, with outputs evaluated using task-specific classification, correlation, regression, and sequence-function metrics \citep{he2024biology}. We use a 500K-example subset that contains all 21 task categories. After Qwen3 tokenization and truncation to 2048 tokens, this subset contains 90,019,678 input tokens, of which 9,003,134 are answer-side supervision tokens. The reported Biology-Instructions score is the average over the 21 task-level metrics on the common 0--100 reporting scale. Biology-Instructions is evaluated deterministically. Full per-task results are provided in Appendix~\ref{sec:appendix-per-task-domain}.

\paragraph{OpenSWI.}
OpenSWI studies surface-wave dispersion-curve inversion \citep{liu2025openswi}. Each input contains period, phase-velocity, and group-velocity values, while the target physical profile contains depth and subsurface properties including P-wave velocity, S-wave velocity, and density. We sample 30K training instances and evaluate the shallow generation-style setting, where models generate a near-surface S-wave velocity profile and performance is measured by RMSE. Because OpenSWI uses stochastic decoding, we report the mean over five runs with decoding seeds 42--46. We exclude the deep setting because its substantially longer and more structurally complex output profiles are not generated reliably under the current autoregressive text formulation.

\paragraph{DISC-Law-SFT and LawBench.}
For legal specialization, we construct the Qwen3-formatted training set from the released DISC-Law-SFT Pair and Triplet files \citep{yue2024lawllm}. The final set contains 55,295 examples after task-balanced sampling, format-targeted supplementation, and removal of overlaps with the LawBench evaluation data. We evaluate with the local OpenCompass port of LawBench in zero-shot mode \citep{fei2024lawbench}. The reported average covers 19 tasks; we exclude ``2-10 trigger\_word\_extraction'' because the adopted evaluator matches semicolon-separated trigger words position by position and is therefore sensitive to the order of an otherwise correct set. Full per-task results and the exclusion rationale are reported in Appendix~\ref{sec:appendix-per-task-domain}.

\subsection{General Capability Evaluation}

We evaluate general-capability retention with the same task configuration for each original backbone and its adapted variants. MATH-500 measures mathematical reasoning with verification accuracy; C-Eval measures Chinese knowledge and reasoning with accuracy; IFEval measures instruction following with prompt-strict accuracy; MMLU-Redux measures broad academic knowledge with accuracy; and INCLUDE evaluates multilingual and cross-cultural knowledge with accuracy \citep{lightman2024let,huang2023c,zhou2023instruction,gema2025we,romanou2025include}. MATH-500, C-Eval, IFEval, and MMLU-Redux are evaluated in thinking mode using lm-evaluation-harness configurations aligned with the Qwen3 Technical Report \citep{yang2025qwen3}. For MATH-500, IFEval, and MMLU-Redux, we report the mean over five runs with decoding seeds 42--46; C-Eval and INCLUDE use deterministic evaluation. The report does not fully specify the evaluation protocol for INCLUDE; we therefore follow its reported parameter settings where available and use the 5-shot multiple-choice likelihood configuration in lm-evaluation-harness without a chat template to approximate the setup as closely as possible. All variants of a given backbone use identical task configurations and metric implementations. The general score is the unweighted average of these five benchmark scores on a 0--100 scale.

\subsection{Router Training Data}

Router training combines domain examples with general examples so that the router learns both when to invoke memory and when to preserve the backbone distribution. The Biology-Instructions, OpenSWI, and legal routers use 2,250, 6,000, and 5,610 domain examples, respectively, together with 15,336 general examples for each router. All general examples are sampled from NVIDIA's Nemotron-Post-Training-Dataset-v1 \citep{bercovich2025llama}.

\section{Datastore and Retrieval-Teacher Construction}
\label{sec:appendix-datastore}

The datastore is built only from answer-side supervision tokens. During preprocessing, SFT examples are represented as a concatenation of query tokens and answer tokens, while labels on query positions are masked out. We use this label mask to construct datastore key-value pairs only for answer positions, so the query affects the hidden state through the prefix context but is not itself stored as a datastore value.

For the feature extractor $\phi$, we use the input hidden state to the MLP of the final decoder block of the frozen datastore encoder. For Qwen models, this corresponds to hooking the final Transformer layer before its MLP sublayer, after the preceding attention and residual computation. In the reported Qwen3 experiments, the KNN teacher is constructed with a Qwen3-8B datastore encoder, yielding 4096-dimensional keys. The Biology-Instructions main experiments and the memory-scale ablation use the same KNN teacher; the memory-scale ablation changes the memory LM size rather than the KNN teacher model. Datastore keys are stored in float16 and cast to float32 for FAISS indexing and search.

We use FAISS L2 nearest-neighbor search to construct the retrieval-based teacher distribution. The keys are not additionally L2-normalized before indexing, so the retrieval metric is not cosine similarity. For each queried answer position, we retrieve the top 2048 neighbors and discard the first nearest neighbor, which corresponds to the query itself when constructing teacher targets from the same datastore. The effective non-self neighbor set therefore contains 2047 entries, and the teacher distribution uses temperature $\tau=16.0$ in Eq.~(\ref{eq:teacher-dist}). Neighbor values with the same token id are merged into a sparse vocabulary distribution, which is then used as the soft target for memory training.

\section{Training and Implementation Details}
\label{sec:appendix-hyperparams}

This section summarizes the training configurations used for the main baselines and MemSFT. Baseline and memory-training runs use cosine learning-rate scheduling with warmup unless otherwise specified, while router training uses a linear schedule. Unless otherwise specified, the maximum sequence length is 2048; the legal SFT and LoRA runs use maximum sequence length 3072. LoRA baselines use an all-linear placement: adapters are applied to the attention projections ($q$, $k$, $v$, $o$) and MLP projections ($gate$, $up$, $down$), and are not applied to embedding layers, normalization layers, or the language-model head.

For full SFT and LoRA baselines, Biology-Instructions models are trained for one epoch on the 500K Biology-Instructions subset, OpenSWI models are trained for three epochs on the 30K OpenSWI subset, and DISC-Law models are trained on the sampled 55K DISC-Law-SFT subset, with three epochs for full SFT and one epoch for LoRA. Full SFT uses the standard answer-side SFT loss. LoRA uses rank 8, alpha 16, dropout 0.05, learning rate $3{\times}10^{-4}$, and the all-linear target-module placement described above.
We use a fixed LoRA hyperparameter configuration across all evaluated backbones for consistency and comparability. We use a 5\% warmup ratio for full SFT and a 1\% warmup ratio for LoRA.

For MemSFT memory training, we train a Qwen3-8B memory on the 500K Biology-Instructions subset for one epoch with learning rate $2{\times}10^{-5}$ and $\beta=0.3$, a Qwen3-8B memory on the 30K OpenSWI subset for three epochs with learning rate $5{\times}10^{-5}$ and $\beta=0.2$, and a Qwen3-8B legal memory on the sampled 55K DISC-Law-SFT subset for three epochs with learning rate $1{\times}10^{-5}$ and $\beta=0.1$. In preliminary runs, we found $\beta$ in the range 0.1--0.3 to produce similar behavior, and therefore use domain-specific values within this range without further tuning. For the main Biology-Instructions experiments, memory training uses 16 A800-80GB GPUs with per-device batch size 4 and gradient accumulation 2, resulting in a global batch size of 128; the Biology-Instructions memory is trained with 150 warmup steps and then reused unchanged across all evaluated Qwen3 backbones.
For the LLaMA2 validation in Table~\ref{tab:llama2-validation}, we train a LLaMA-family 7B memory on the same 500K Biology-Instructions subset and use the same memory-router training pipeline as in the Qwen3 Biology-Instructions experiments.

For router training, both the backbone LM and the memory LM are frozen, and only the token-level router is updated. In the main configuration, the router uses both branches' hidden states together with scalar uncertainty features derived from their output distributions. Specifically, the scalar inputs include base confidence, base entropy, memory confidence, and memory entropy. These scalar features are projected to a 64-dimensional representation and combined with the hidden-state features before being passed to a two-layer MLP with hidden dimension 128. The router outputs two log-weights, $\log(1-\lambda_t)$ and $\log(\lambda_t)$, which are used for distribution-level fusion of the backbone and memory distributions. Across domains and backbone sizes, router training uses one epoch, AdamW with a linear learning-rate schedule, per-device batch size 1, gradient accumulation 8, maximum sequence length 2048, and 100 warmup optimizer updates; the learning rate is adjusted for each backbone size. We set $\alpha_s=0.2$, with $s_t=-32$ for domain examples and $s_t=+32$ for general examples, and average the router objective over all answer-side supervised tokens in each microbatch. The resulting router remains small relative to either LM branch: the measured Biology-Instructions routers contain 1,082,242 trainable parameters for the Qwen3-8B backbone and 1,213,314 parameters for the Qwen3-14B and Qwen3-32B backbones.

For the memory-scale analysis in Figure~\ref{fig:memory-scale}, the 1.7B, 4B, and 8B memories use the same 500K Biology-Instructions training data, Qwen3-8B retrieval teacher, frozen Qwen3-14B backbone, and overall memory-router training pipeline as the main setting in Table~\ref{tab:bioins-main}. Memory size is the architectural variable, with size-specific optimization configurations used for each memory. Specifically, the 1.7B and 4B memories use learning rates of $1{\times}10^{-4}$ and $5{\times}10^{-5}$, respectively, while retaining the same one-epoch memory-training recipe; each memory size uses a separately trained compatible router.

\subsection{Additional Router Case Studies}
\label{sec:appendix-router-cases}

Figures~\ref{fig:router-openswi-case}--\ref{fig:appendix-mmlu-mars-greenhouse-router-case} analyze router weights from OpenSWI to BioIns, MATH, and MMLU. Each box gives the question and output, panel (a) shows token-level weights, and panel (b) gives their role-wise distribution; the cases progress from domain-dependent to general tasks.

\Needspace{4\baselineskip}
\subsubsection{BioIns Case}
\begin{figure*}[!t]
    \centering
    \begin{minipage}{0.965\textwidth}
    \begin{tcolorbox}[
        colback=blue!2,
        colframe=blue!35!white,
        boxrule=0.8pt,
        arc=10pt,
        outer arc=10pt,
        left=8pt,
        right=8pt,
        top=6pt,
        bottom=7pt
    ]
        \small
        \noindent Answer the question:\hfill
        \colorbox{blue!9}{\strut\textbf{BioIns Case}}

        \vspace{0.35em}
        \noindent\textbf{Question:} For core-promoter detection, is the following DNA sequence a core promoter?
        \par\vspace{0.1em}\noindent\texttt{\footnotesize <DNA>GAATGAACAGGAGTCGGTTCTCACCCAACTTCCATTAAGGACTCGGGGCAGGAGGGGCAGAAGTTGCGCG</DNA>}

        \vspace{0.2em}
        \noindent\tikz{\draw[gray!65,dashed,line width=0.5pt] (0,0)--(\linewidth,0);}

        \vspace{0.35em}
        \noindent\textbf{Output:} The sequence contains core-promoter-like motifs. Therefore, the final answer is $\boxed{\mathrm{positive}}$.
    \end{tcolorbox}
    \end{minipage}

    \vspace{0.7em}
    \begin{minipage}[c]{0.600\textwidth}
        \centering
        \includegraphics[width=\linewidth]{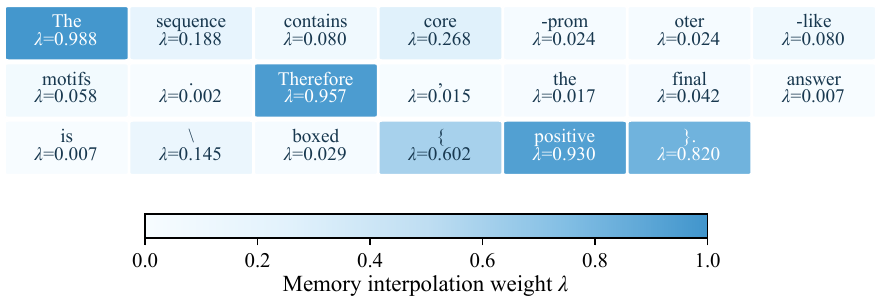}
        \vspace{-0.4em}

        \textbf{(a)} Detailed demonstration of router-assigned lambda.
    \end{minipage}\hfill
    \begin{minipage}[c]{0.370\textwidth}
        \centering
        \includegraphics[width=\linewidth]{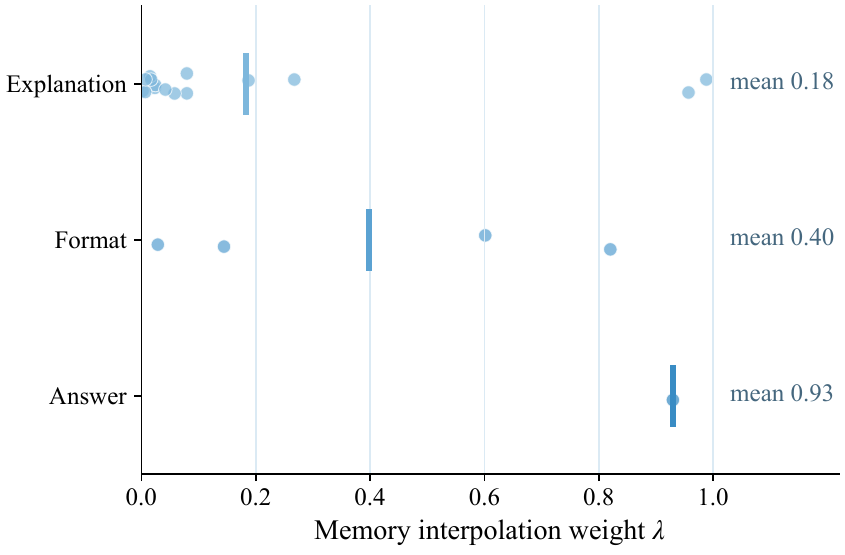}
        \vspace{0.2em}

        \textbf{(b)} Lambda distribution.
    \end{minipage}
    \caption{Router behavior on a BioIns core-promoter classification case. Panel (a) displays all 20 answer tokens. Panel (b) groups them into explanation, boxed-format, and answer tokens; vertical bars mark category means.}
    \label{fig:appendix-bioins-router-case}
\end{figure*}

Figure~\ref{fig:appendix-bioins-router-case} shows mean $\lambda$ rising from 0.18 for explanation tokens to 0.40 for boxed-format tokens and 0.93 for \texttt{positive}, concentrating memory on the specialized BioIns decision.

\FloatBarrier
\Needspace{4\baselineskip}
\subsubsection{MATH Case I: Alternating Sum}
\begin{figure*}[!t]
    \centering
    \begin{minipage}{0.965\textwidth}
    \begin{tcolorbox}[
        colback=cyan!3,
        colframe=cyan!45!green,
        boxrule=0.8pt,
        arc=10pt,
        outer arc=10pt,
        left=8pt,
        right=8pt,
        top=6pt,
        bottom=7pt
    ]
        \small
        \noindent Answer the question:\hfill
        \colorbox{cyan!10}{\strut\textbf{MATH Case I}}

        \vspace{0.35em}
        \noindent\textbf{Question:} Compute $1-2+3-4+5-\dots+99-100$.

        \vspace{0.2em}
        \noindent\tikz{\draw[gray!65,dashed,line width=0.5pt] (0,0)--(\linewidth,0);}

        \vspace{0.35em}
        \noindent\textbf{Output:} Group the terms in pairs. Each pair is of the form $(2k-1)-2k=-1$. Since the series ends at 100 (an even number), the total number of such pairs is: $50\ \text{pairs}$ Each pair evaluates to $-1$, so the total sum is: $50\times(-1)=-50$ Final Answer $\boxed{-50}$
    \end{tcolorbox}
    \end{minipage}

    \vspace{0.7em}
    \begin{minipage}[c]{0.600\textwidth}
        \centering
        \includegraphics[width=\linewidth]{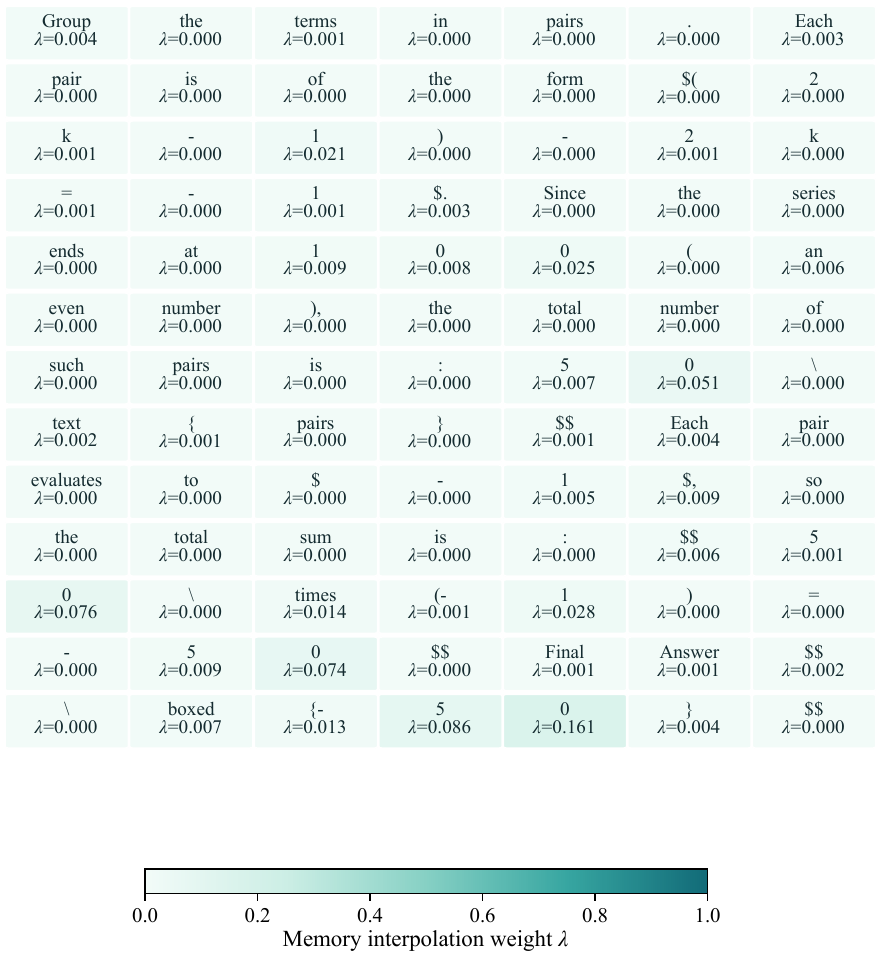}
        \vspace{-0.4em}

        \textbf{(a)} Detailed demonstration of router-assigned lambda.
    \end{minipage}\hfill
    \begin{minipage}[c]{0.370\textwidth}
        \centering
        \includegraphics[width=\linewidth]{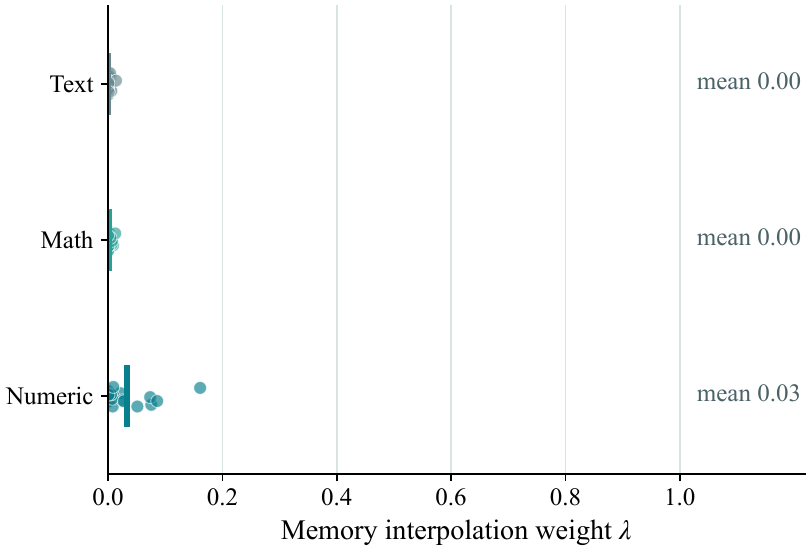}
        \vspace{0.2em}

        \textbf{(b)} Lambda distribution.
    \end{minipage}
    \caption{Router behavior on the alternating-sum MATH case. Panel (a) follows the displayed derivation from pairing consecutive terms to the boxed result. Panel (b) groups the same displayed tokens into text, mathematical-formatting, and numeric categories.}
    \label{fig:appendix-math-pair-router-case}
\end{figure*}

Figure~\ref{fig:appendix-math-pair-router-case} contrasts the specialized BioIns answer with a general arithmetic derivation that stays backbone-dominant. Among the displayed tokens, text and mathematical-formatting tokens have mean weights of 0.001 and 0.002, respectively, while numeric tokens average 0.033. The increase for numbers is measurable but remains far below the near-saturated routing of OpenSWI profile values.

\FloatBarrier
\Needspace{4\baselineskip}
\subsubsection{MATH Case II: Complex Rotation}
\begin{figure*}[!t]
    \centering
    \begin{minipage}{0.965\textwidth}
    \begin{tcolorbox}[
        colback=cyan!3,
        colframe=cyan!45!green,
        boxrule=0.8pt,
        arc=10pt,
        outer arc=10pt,
        left=8pt,
        right=8pt,
        top=6pt,
        bottom=7pt
    ]
        \small
        \noindent Answer the question:\hfill
        \colorbox{cyan!10}{\strut\textbf{MATH Case II}}

        \vspace{0.35em}
        \noindent\textbf{Question:} Let $z=2+\sqrt{2}-(3+3\sqrt{2})i$ and $c=2-3i$. Let $w$ be the result when $z$ is rotated around $c$ counter-clockwise by $\pi/4$. Find $w$.

        \vspace{0.2em}
        \noindent\tikz{\draw[gray!65,dashed,line width=0.5pt] (0,0)--(\linewidth,0);}

        \vspace{0.35em}
        \noindent\textbf{Output:} \textbf{Step 1: Translate the point $z$ by subtracting $c$.} Move the center of rotation to the origin:
        $z-c=\sqrt{2}-3\sqrt{2}i$.
        \textbf{Step 2: Apply the rotation by $\pi/4$.}
        $(1-3i)(1+i)=1+i-3i-3i^2=1-2i+3=4-2i$.
        \textbf{Step 3: Translate back by adding $c$.}
        $w=(4-2i)+(2-3i)=6-5i$. \textbf{Final Answer:} $\boxed{6-5i}$.
    \end{tcolorbox}
    \end{minipage}

    \vspace{0.7em}
    \begin{minipage}[c]{0.600\textwidth}
        \centering
        \includegraphics[width=\linewidth]{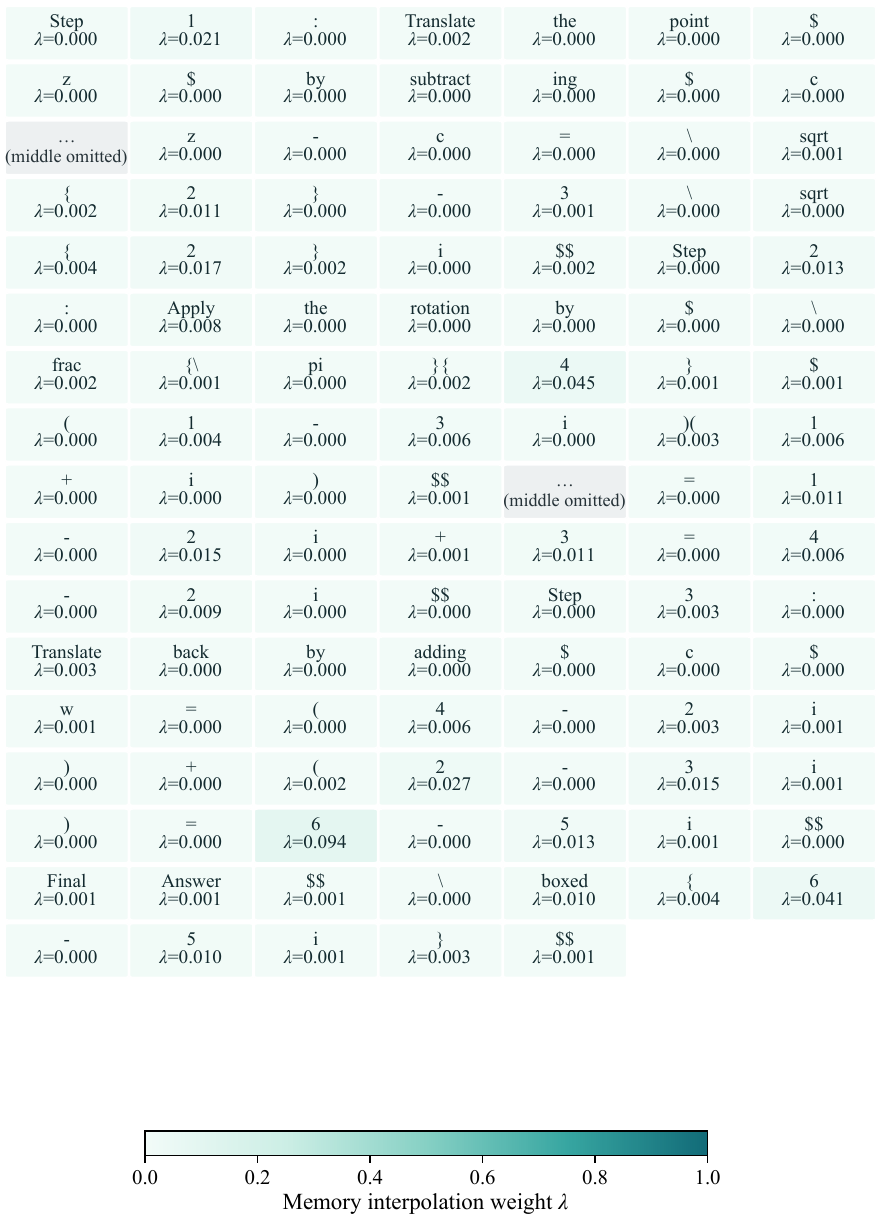}
        \vspace{-0.4em}

        \textbf{(a)} Detailed demonstration of router-assigned lambda.
    \end{minipage}\hfill
    \begin{minipage}[c]{0.370\textwidth}
        \centering
        \includegraphics[width=\linewidth]{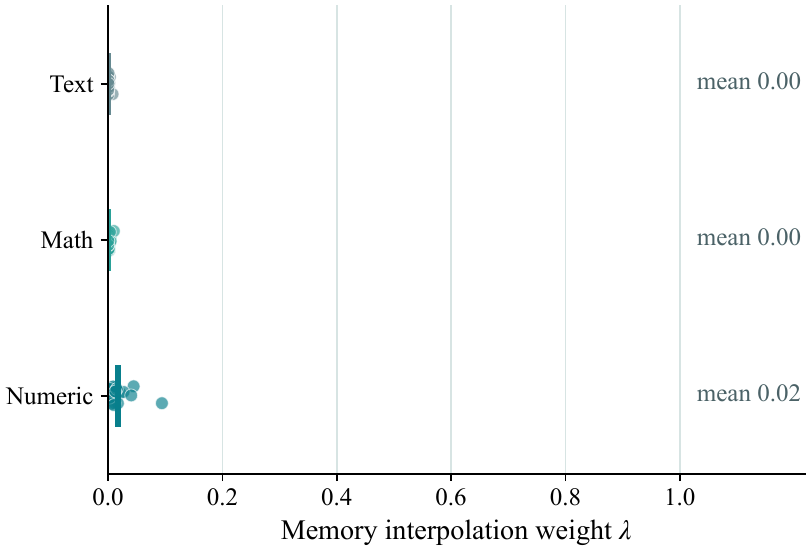}
        \vspace{0.2em}

        \textbf{(b)} Lambda distribution.
    \end{minipage}
    \caption{Router behavior on a complex-rotation MATH case. Panel (a) traces the three-step solution, with intermediate explanatory and algebraic tokens omitted for compactness. Panel (b) groups the displayed tokens into text, mathematical-formatting, and numeric categories.}
    \label{fig:appendix-math-complex-rotation-router-case}
\end{figure*}

Figure~\ref{fig:appendix-math-complex-rotation-router-case} is a stricter low-routing control. Across the displayed excerpts of the three-step calculation, text and mathematical-formatting tokens have mean weights of 0.0006 and 0.0009, respectively, while numeric tokens average 0.0169. Every displayed token remains below 0.1 (maximum 0.0942), showing that even a multi-step symbolic calculation can remain almost entirely backbone-routed.

\FloatBarrier
\Needspace{4\baselineskip}
\subsubsection{MMLU Case I: Narcolepsy}
\begin{figure*}[!t]
    \centering
    \begin{minipage}{0.965\textwidth}
    \begin{tcolorbox}[
        colback=casepurple!3!white,
        colframe=casepurple!45!white,
        boxrule=0.8pt,
        arc=10pt,
        outer arc=10pt,
        left=8pt,
        right=8pt,
        top=6pt,
        bottom=7pt
    ]
        \small
        \noindent Answer the question:\hfill
        \colorbox{casepurple!11!white}{\strut\textbf{MMLU Case I}}

        \vspace{0.35em}
        \noindent\textbf{Question:} Which disorder is characterized by uncontrollable episodes of falling asleep during the day? \textbf{A.} Dyslexia; \textbf{B.} Epilepsy; \textbf{C.} Hydrocephalus; \textbf{D.} Narcolepsy.

        \vspace{0.2em}
        \noindent\tikz{\draw[gray!65,dashed,line width=0.5pt] (0,0)--(\linewidth,0);}

        \vspace{0.35em}
        \noindent\textbf{Output:} I remember that the key symptom here is the uncontrollable episodes of falling asleep during the day, which is the hallmark of narcolepsy. The other options don't fit this description. So the correct answer should be D. Narcolepsy. D
    \end{tcolorbox}
    \end{minipage}

    \vspace{0.7em}
    \begin{minipage}[c]{0.600\textwidth}
        \centering
        \includegraphics[width=\linewidth]{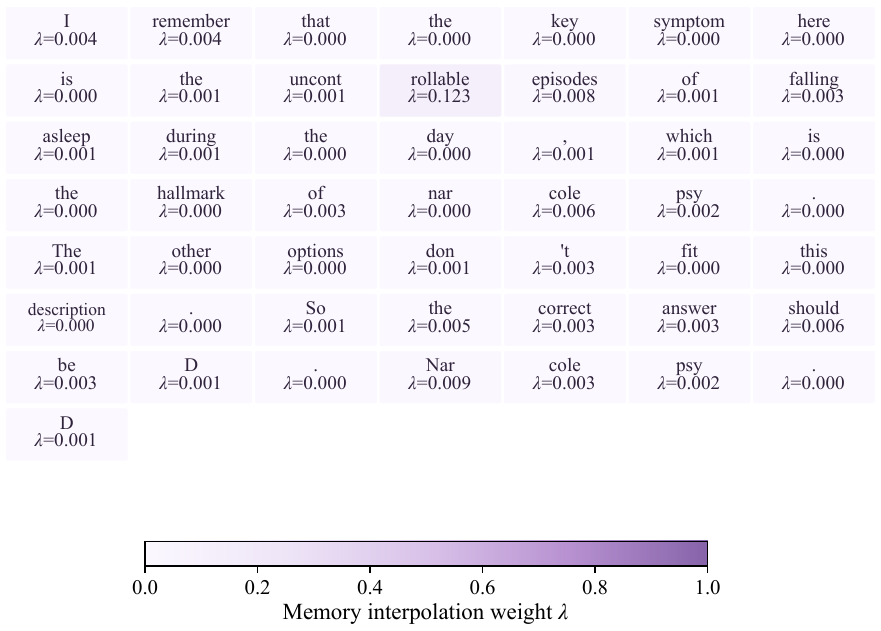}
        \vspace{-0.4em}

        \textbf{(a)} Detailed demonstration of router-assigned lambda.
    \end{minipage}\hfill
    \begin{minipage}[c]{0.370\textwidth}
        \centering
        \includegraphics[width=\linewidth]{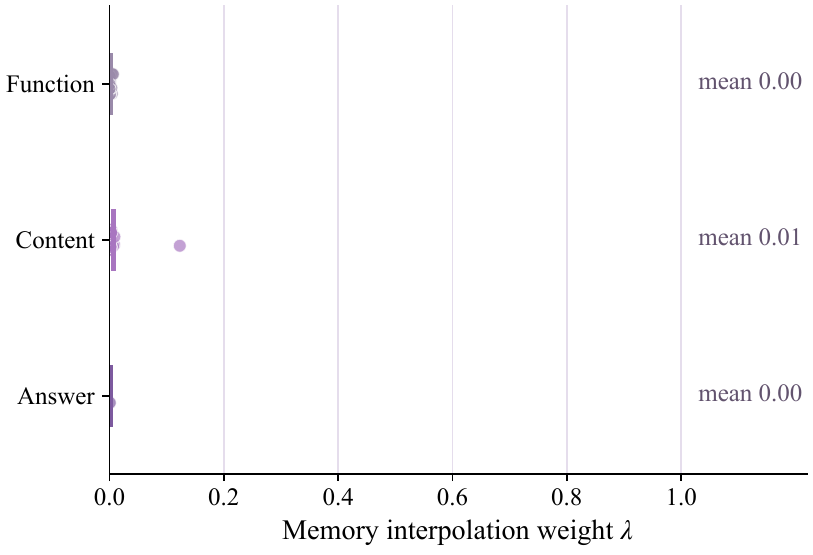}
        \vspace{0.2em}

        \textbf{(b)} Lambda distribution.
    \end{minipage}
    \caption{Router behavior on an MMLU anatomy question about narcolepsy. Panel (a) shows the concluding analysis and final option token. Panel (b) groups exactly the displayed sequence into function, content, and answer tokens.}
    \label{fig:appendix-mmlu-narcolepsy-router-case}
\end{figure*}

Figure~\ref{fig:appendix-mmlu-narcolepsy-router-case} considers a general factual-recognition question rather than a BioIns-memory task. Within the displayed sequence, function and content tokens have mean weights of 0.0012 and 0.0070, respectively, and the final \texttt{D} token has weight 0.0006. The router therefore leaves both the explanation and the decision to the frozen backbone.

\FloatBarrier
\Needspace{4\baselineskip}
\subsubsection{MMLU Case II: Mars Greenhouse Effect}
\begin{figure*}[!t]
    \centering
    \begin{minipage}{0.965\textwidth}
    \begin{tcolorbox}[
        colback=casepurple!3!white,
        colframe=casepurple!45!white,
        boxrule=0.8pt,
        arc=10pt,
        outer arc=10pt,
        left=8pt,
        right=8pt,
        top=6pt,
        bottom=7pt
    ]
        \small
        \noindent Answer the question:\hfill
        \colorbox{casepurple!11!white}{\strut\textbf{MMLU Case II}}

        \vspace{0.35em}
        \noindent\textbf{Question:} Mars has an atmosphere that is almost entirely carbon dioxide. Why is there not a strong greenhouse effect keeping the planet warm? \textbf{A.} Mars does not have enough internal heat to drive the greenhouse effect; \textbf{B.} Mars is too far from the Sun for the greenhouse effect to work; \textbf{C.} the greenhouse effect requires an ozone layer, which Mars does not have; \textbf{D.} the atmosphere on Mars is too thin to trap a significant amount of heat.

        \vspace{0.2em}
        \noindent\tikz{\draw[gray!65,dashed,line width=0.5pt] (0,0)--(\linewidth,0);}

        \vspace{0.35em}
        \noindent\textbf{Output:} Option D states the atmosphere is too thin. Mars' atmosphere is very thin compared to Venus. Even though CO$_2$ is a greenhouse gas, if the atmosphere is too thin, it can't trap much heat. That makes sense. Venus has a dense atmosphere, so even though it's CO$_2$, it's effective. Mars' thin atmosphere can't trap heat as effectively, so D seems correct. D
    \end{tcolorbox}
    \end{minipage}

    \vspace{0.7em}
    \begin{minipage}[c]{0.600\textwidth}
        \centering
        \includegraphics[width=\linewidth]{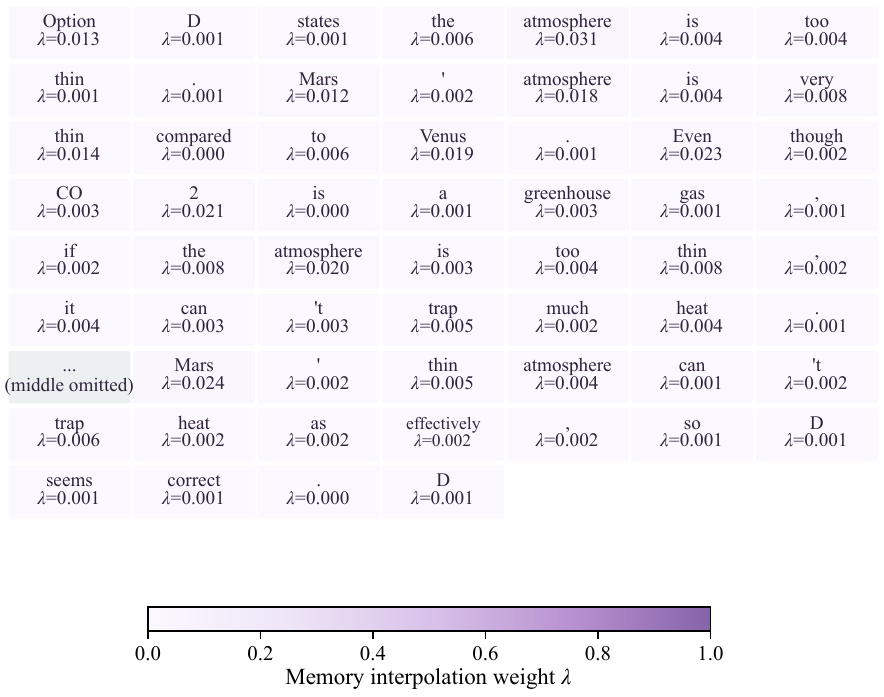}
        \vspace{-0.4em}

        \textbf{(a)} Detailed demonstration of router-assigned lambda.
    \end{minipage}\hfill
    \begin{minipage}[c]{0.370\textwidth}
        \centering
        \includegraphics[width=\linewidth]{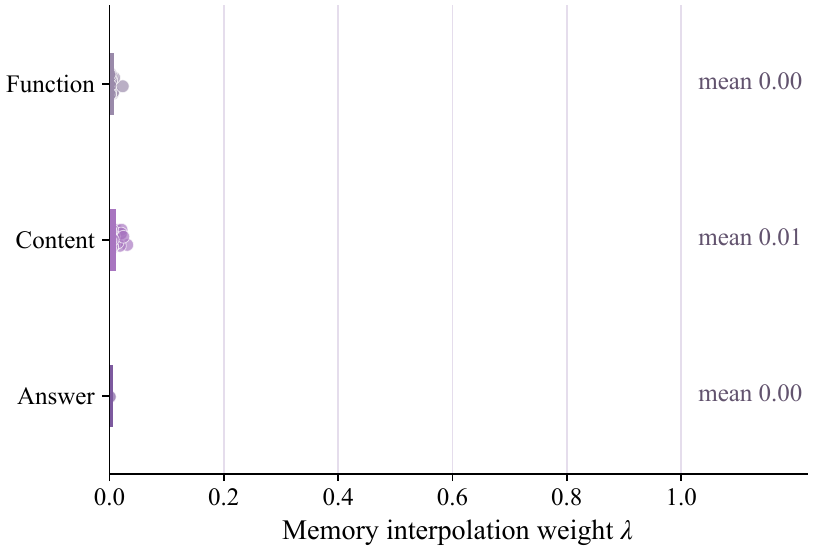}
        \vspace{0.2em}

        \textbf{(b)} Lambda distribution.
    \end{minipage}
    \caption{Router behavior on an MMLU astronomy question about Mars' greenhouse effect. Panel (a) displays the retained tokens from the output, with a middle span omitted for compactness. Panel (b) groups the output tokens into function, content, and answer categories.}
    \label{fig:appendix-mmlu-mars-greenhouse-router-case}
\end{figure*}

Figure~\ref{fig:appendix-mmlu-mars-greenhouse-router-case} shows a second uniformly backbone-dominant MMLU example. Across the complete displayed answer, function and content tokens have mean weights of 0.0027 and 0.0066, and the final \texttt{D} token has weight 0.0005; the complete reasoning-and-answer trace also remains below 0.1 (maximum 0.0591). Together with Figure~\ref{fig:router-openswi-case} and the other appendix cases, this comparison indicates that memory use follows task relevance: it is strongest for specialized biological labels and geophysical profile values, remains low for ordinary mathematical reasoning, and is negligible for general multiple-choice answers.

\FloatBarrier

\section{Full Per-Task Domain Results}
\label{sec:appendix-per-task-domain}

The main tables report average domain performance for compactness. Here we provide the corresponding per-task domain results for Biology-Instructions and LawBench. For Biology-Instructions, Table~\ref{tab:appendix-bioins-task-results-small} expands the Qwen3-8B and Qwen3-14B settings, while Table~\ref{tab:appendix-bioins-task-results} expands the representative Qwen3-32B setting and the largest Qwen3-235B-A22B setting. For LawBench, Table~\ref{tab:appendix-lawbench-task-results} reports the 19 evaluated subtasks.

\subsection{Biology-Instructions}

\begin{table}[H]
\centering
\scriptsize
\setlength{\tabcolsep}{2.1pt}
\resizebox{\textwidth}{!}{%
\begin{tabular}{llrrrrrrrr}
\toprule
Task & Metric & Qwen3-8B & 8B+SFT & 8B+LoRA & 8B+MemSFT & Qwen3-14B & 14B+SFT & 14B+LoRA & 14B+MemSFT \\
\midrule
DNA-cpd & MCC & -10.45 & 60.18 & 57.38 & 62.14 & -0.69 & 60.90 & 58.67 & 62.74 \\
DNA-emp & MCC & -1.26 & 12.44 & 11.37 & 15.04 & 0.20 & 16.65 & 17.80 & 16.44 \\
DNA-pd & MCC & -13.20 & 75.12 & 70.87 & 74.61 & 0.22 & 75.80 & 76.60 & 74.81 \\
DNA-tf-h & MCC & 0.47 & 30.27 & 13.76 & 41.80 & -0.72 & 27.20 & 21.46 & 40.48 \\
DNA-tf-m & MCC & -0.18 & 24.86 & 37.38 & 36.10 & 0.28 & 12.50 & 26.43 & 33.86 \\
Multi\_sequence-antibody\_antigen & MCC & 2.00 & 9.51 & -0.51 & 25.55 & -6.04 & 30.91 & 2.01 & 25.33 \\
Multi\_sequence-promoter\_enhancer\_interaction & MCC & -12.85 & 0.84 & -6.08 & -5.30 & -0.64 & 0.19 & -0.10 & -3.59 \\
Multi\_sequence-rna\_protein\_interaction & MCC & -2.72 & 74.91 & 48.69 & 78.52 & -2.89 & 75.50 & 55.87 & 78.92 \\
DNA-enhancer\_activity & PCC & 0.58 & 40.45 & 32.95 & 38.84 & 0.15 & 41.13 & 34.27 & 42.25 \\
RNA-CRISPROnTarget & Spearman & -3.67 & 4.44 & 10.52 & 17.97 & 1.06 & 10.02 & 2.94 & 16.14 \\
Protein-Fluorescence & Spearman & 4.59 & 0.00 & 2.90 & 0.00 & -6.09 & 0.00 & -1.25 & 0.00 \\
Protein-Stability & Spearman & 1.94 & 59.01 & 44.22 & 49.75 & 1.18 & 57.44 & 6.23 & 47.90 \\
Protein-Thermostability & Spearman & -0.53 & 51.79 & 51.98 & 53.29 & -2.36 & 52.47 & 51.17 & 53.59 \\
RNA-Isoform & $R^2$ & 0.39 & 82.56 & 79.59 & 85.49 & 0.17 & 82.68 & 82.79 & 85.31 \\
RNA-MeanRibosomeLoading & $R^2$ & 0.06 & 11.70 & 4.41 & 58.46 & 0.74 & 50.61 & 1.68 & 58.86 \\
RNA-ProgrammableRNASwitches & $R^2$ & 0.11 & 22.27 & 15.44 & 22.42 & 0.09 & 21.22 & 21.10 & 16.93 \\
RNA-Modification & AUC & 51.75 & 52.47 & 51.55 & 53.88 & 50.67 & 54.91 & 53.49 & 54.34 \\
Protein-Solubility & Acc. & 46.50 & 64.50 & 61.60 & 68.90 & 51.50 & 61.10 & 59.10 & 67.70 \\
RNA-NoncodingRNAFamily & Acc. & 8.20 & 43.70 & 27.80 & 64.00 & 15.90 & 64.30 & 45.40 & 64.70 \\
Protein-FunctionEC & Score & 1.36 & 10.76 & 6.59 & 7.09 & 2.93 & 14.46 & 9.74 & 13.42 \\
Multi\_sequence-sirnaEfficiency & Mixed & 33.41 & 52.27 & 51.01 & 51.13 & 33.85 & 54.36 & 48.15 & 51.24 \\
\midrule
Biology-Instructions AVG & -- & 5.07 & 37.34 & 32.07 & 42.84 & 6.64 & 41.16 & 32.07 & 42.92 \\
\bottomrule
\end{tabular}
}
\caption{Full per-task Biology-Instructions results for Qwen3-8B and Qwen3-14B. Metrics follow the Biology-Instructions evaluation protocol, and all values are reported on the same scale as the main Biology-Instructions table, where higher is better.}
\label{tab:appendix-bioins-task-results-small}
\end{table}

\begin{table}[H]
\centering
\scriptsize
\setlength{\tabcolsep}{2.5pt}
\resizebox{\textwidth}{!}{%
\begin{tabular}{llrrrrrr}
\toprule
Task & Metric & Qwen3-32B & 32B+SFT & 32B+LoRA & 32B+MemSFT & Qwen3-235B-A22B & 235B-A22B+MemSFT \\
\midrule
DNA-cpd & MCC & 3.06 & 59.30 & 59.77 & 61.95 & -15.20 & 62.90 \\
DNA-emp & MCC & 3.85 & 18.03 & 18.35 & 16.62 & -13.47 & 18.17 \\
DNA-pd & MCC & -6.56 & 78.02 & 74.52 & 75.62 & -26.10 & 74.46 \\
DNA-tf-h & MCC & -3.91 & 40.59 & 37.84 & 41.71 & 1.13 & 40.93 \\
DNA-tf-m & MCC & -5.29 & 42.66 & 37.89 & 34.31 & 1.39 & 32.87 \\
Multi\_sequence-antibody\_antigen & MCC & -11.59 & 11.93 & 26.49 & 24.42 & -16.78 & 23.13 \\
Multi\_sequence-promoter\_enhancer\_interaction & MCC & 0.30 & 6.40 & 5.81 & -5.94 & -21.85 & -4.52 \\
Multi\_sequence-rna\_protein\_interaction & MCC & 0.30 & 79.61 & 77.92 & 78.52 & -11.20 & 77.92 \\
DNA-enhancer\_activity & PCC & -0.95 & 41.13 & 37.69 & 40.15 & 1.37 & 41.03 \\
RNA-CRISPROnTarget & Spearman & 1.37 & 12.13 & 13.07 & 17.72 & 0.09 & 16.77 \\
Protein-Fluorescence & Spearman & 3.19 & 0.00 & 2.08 & -4.03 & -1.48 & 0.00 \\
Protein-Stability & Spearman & 2.16 & 47.32 & 41.33 & 53.25 & -2.50 & 51.24 \\
Protein-Thermostability & Spearman & 2.73 & 54.81 & 49.19 & 54.57 & 0.87 & 53.17 \\
RNA-Isoform & $R^2$ & 0.23 & 82.26 & 82.40 & 85.21 & 0.11 & 84.38 \\
RNA-MeanRibosomeLoading & $R^2$ & 1.13 & 60.37 & 43.45 & 59.52 & 0.00 & 49.61 \\
RNA-ProgrammableRNASwitches & $R^2$ & 0.06 & 20.78 & 16.33 & 15.54 & 0.08 & 16.12 \\
RNA-Modification & AUC & 50.61 & 53.96 & 53.35 & 55.26 & 50.50 & 55.78 \\
Protein-Solubility & Acc. & 48.70 & 69.20 & 66.00 & 70.00 & 40.60 & 70.00 \\
RNA-NoncodingRNAFamily & Acc. & 4.80 & 71.00 & 62.20 & 60.00 & 0.20 & 58.40 \\
Protein-FunctionEC & Score & 3.24 & 16.46 & 11.80 & 13.52 & 3.21 & 9.05 \\
Multi\_sequence-sirnaEfficiency & Mixed & 33.83 & 52.73 & 51.01 & 51.31 & 32.97 & 51.56 \\
\midrule
Biology-Instructions AVG & -- & 6.25 & 43.75 & 41.36 & 42.82 & 1.14 & 42.05 \\
\bottomrule
\end{tabular}
}
\caption{Full per-task Biology-Instructions results for the representative Qwen3-32B setting and the largest Qwen3-235B-A22B setting. Metrics follow the Biology-Instructions evaluation protocol, and all values are reported on the same scale as the main Biology-Instructions table, where higher is better.}
\label{tab:appendix-bioins-task-results}
\end{table}

\subsection{LawBench}

Following our evaluation configuration, LawBench ``2-10 trigger\_word\_extraction'' is excluded from the aggregate. This task is naturally set-like, but the current evaluator scores trigger words position by position after splitting the output string, making the score sensitive to the order of otherwise correct trigger words. We therefore exclude it from the main comparable LawBench average unless all systems are re-evaluated under a revised set-based metric.

\begin{table}[H]
\centering
\scriptsize
\setlength{\tabcolsep}{3pt}
\resizebox{\textwidth}{!}{%
\begin{tabular}{lrrrrr}
\toprule
LawBench Task & Qwen3-14B & SFT & LoRA & MemSFT & $\Delta$ vs. Qwen3-14B \\
\midrule
1-1 article\_recitation & 22.53 & 21.58 & 27.62 & 24.63 & +2.10 \\
1-2 knowledge\_question\_answering & 56.00 & 73.60 & 79.00 & 72.60 & +16.60 \\
2-1 document\_proofreading & 30.80 & 55.34 & 62.87 & 23.69 & -7.11 \\
2-2 dispute\_focus\_identification & 36.00 & 36.80 & 38.80 & 45.60 & +9.60 \\
2-3 marital\_disputes\_identification & 62.13 & 71.95 & 63.11 & 69.34 & +7.21 \\
2-4 issue\_topic\_identification & 39.80 & 43.20 & 40.80 & 44.20 & +4.40 \\
2-5 reading\_comprehension & 54.82 & 35.49 & 36.91 & 54.76 & -0.06 \\
2-6 named\_entity\_recognition & 7.03 & 41.51 & 41.06 & 71.09 & +64.06 \\
2-7 opinion\_summarization & 31.54 & 32.04 & 35.48 & 30.35 & -1.19 \\
2-8 argument\_mining & 42.00 & 55.20 & 59.80 & 54.80 & +12.80 \\
2-9 event\_detection & 75.80 & 89.79 & 83.25 & 82.09 & +6.29 \\
3-1 fact\_based\_article\_prediction & 83.03 & 81.40 & 85.20 & 79.56 & -3.47 \\
3-2 scene\_based\_article\_prediction & 28.82 & 36.13 & 33.36 & 29.31 & +0.49 \\
3-3 charge\_prediction & 58.32 & 61.62 & 59.99 & 56.35 & -1.97 \\
3-4 prison\_term\_prediction\_wo\_article & 80.09 & 82.86 & 82.41 & 87.06 & +6.97 \\
3-5 prison\_term\_prediction\_w\_article & 80.17 & 82.91 & 82.76 & 86.62 & +6.45 \\
3-6 case\_analysis & 45.00 & 65.20 & 69.20 & 62.40 & +17.40 \\
3-7 criminal\_damages\_calculation & 95.60 & 62.80 & 74.60 & 78.40 & -17.20 \\
3-8 consultation & 17.29 & 16.10 & 16.30 & 20.11 & +2.82 \\
\midrule
LawBench AVG & 49.83 & 55.03 & 56.45 & 56.47 & +6.64 \\
\bottomrule
\end{tabular}
}
\caption{Full per-task LawBench results. The average is computed over the 19 evaluated subtasks and excludes ``2-10 trigger\_word\_extraction'', whose current evaluator is order-sensitive for set-like trigger-word outputs. All values are reported on a 0--100 scale, where higher is better.}
\label{tab:appendix-lawbench-task-results}
\end{table}

\section{Ablating Routing and External Memory}
\label{sec:appendix-routing-external-memory}

We conduct a unified ablation to examine how the routing mechanism and the training strategy for the external module contribute to the overall MemSFT framework. We first keep the frozen Qwen3-14B backbone and Qwen3-8B memory unchanged, but replace learned token-level routing with fixed interpolation coefficients. We then compare two routed systems with different external modules: a Qwen3-8B model trained by standard domain SFT and a Qwen3-8B memory trained with retrieval-based supervision. A separate router is trained for each external module using the same architecture and training recipe. We evaluate all configurations on the official Biology-Instructions benchmark and the five general benchmarks used in the main experiments.

\begin{table}[!t]
\centering
\scriptsize
\setlength{\tabcolsep}{2.2pt}
\resizebox{\textwidth}{!}{%
\begin{tabular}{llccccccc}
\toprule
Routing / Fusion & External Module & BioIns Avg. & MATH-500 & C-Eval & IFEval & MMLU-Redux & INCLUDE & General Avg. \\
\midrule
Backbone only & -- & 6.64 & 96.44 & 83.10 & 85.66 & 90.01 & 60.90 & 83.22 \\
\midrule
Fixed $\lambda=0.1$ & Qwen3-8B memory & 6.28 & 96.32 & 82.84 & 84.40 & 89.77 & 62.74 & 83.21 \\
Fixed $\lambda=0.2$ & Qwen3-8B memory & 6.66 & 94.40 & 82.91 & 74.09 & 82.22 & 62.85 & 79.29 \\
Fixed $\lambda=0.3$ & Qwen3-8B memory & 6.94 & 83.56 & 82.99 & 58.41 & 67.15 & 62.62 & 70.95 \\
Fixed $\lambda=0.5$ & Qwen3-8B memory & 10.51 & 49.68 & 81.87 & 23.40 & 61.17 & 59.78 & 55.18 \\
Fixed $\lambda=0.7$ & Qwen3-8B memory & 29.65 & 22.04 & 73.85 & 14.68 & 60.14 & 51.01 & 44.34 \\
\midrule
Learned router & Qwen3-8B SFT & 37.34 & 96.72 & 83.06 & 86.14 & 90.32 & 62.80 & 83.81 \\
\textbf{Learned router} & \textbf{Qwen3-8B memory} & \textbf{42.92} & 96.48 & 82.99 & 85.40 & 90.49 & 62.77 & 83.62 \\
\bottomrule
\end{tabular}
}
\caption{Ablations of routing and external memory with a frozen Qwen3-14B backbone. The fixed $\lambda$ rows use the same Qwen3-8B memory without a learned router. The final two rows compare a standard domain SFT model with a memory trained using retrieval-based supervision under learned token-level routing. General Avg. is averaged over MATH-500, C-Eval, IFEval, MMLU-Redux, and INCLUDE. All scores are reported on a 0--100 scale.}
\label{tab:appendix-routing-external-memory}
\end{table}

Table~\ref{tab:appendix-routing-external-memory} shows that fixed interpolation cannot effectively balance domain specialization and general capability: increasing the fixed memory weight strengthens domain performance but increasingly interferes with the general benchmarks. Learned token-level routing instead allows the memory to contribute selectively while preserving the behavior of the frozen backbone. Under learned routing, the memory trained with retrieval-based supervision also provides stronger domain specialization than the standard SFT external module, with comparable general performance. Together, these results show how the two components work within the complete MemSFT framework: retrieval-based supervision strengthens the external domain expert, while the router controls how that expertise is integrated with the backbone.

\section{Forgetting-Mitigation Baselines}
\label{sec:appendix-antiforgetting}

Table~\ref{tab:appendix-antiforgetting} reports the full values underlying Figure~\ref{fig:bioins-antiforgetting}. Both baselines for mitigating forgetting use the Qwen3-14B Biology-Instructions setting. MixTraining(1:1) follows the same LoRA configuration as the main Qwen3-14B Biology-Instructions LoRA baseline, but trains on a 1:1 mixture of 500K Biology-Instructions examples and 500K general instruction examples sampled from NVIDIA's Nemotron-Post-Training-Dataset-v1 \citep{bercovich2025llama}. For Wise-FT-style interpolation, we do not train a separate model; instead, we scale the Biology-Instructions LoRA update before merging it into the original backbone, i.e., $\theta=\theta_{\mathrm{base}}+\gamma\Delta_{\mathrm{LoRA}}$, with $\gamma \in \{0.2,0.4,0.6,0.8\}$.

\begin{table}[H]
\centering
\footnotesize
\setlength{\tabcolsep}{4pt}
\resizebox{\textwidth}{!}{%
\begin{tabular}{lccccccc}
\toprule
Method & Biology-Instructions Avg. & MATH-500 & C-Eval & IFEval & MMLU-Redux & INCLUDE & General Avg. \\
\midrule
Qwen3-14B & 6.64 & 96.44 & 83.10 & 85.66 & 90.01 & 60.90 & 83.22 \\
LoRA r8/lr3e-4 & 32.07 & 96.04 & 81.72 & 43.99 & 88.65 & 57.66 & 73.61 \\
Wise-FT $\gamma=0.2$ & 8.25 & 96.52 & 82.91 & 85.77 & 89.88 & 61.00 & 83.22 \\
Wise-FT $\gamma=0.4$ & 19.05 & 96.52 & 82.99 & 85.40 & 89.78 & 60.74 & 83.09 \\
Wise-FT $\gamma=0.6$ & 27.39 & 96.28 & 82.39 & 82.62 & 89.47 & 59.97 & 82.15 \\
Wise-FT $\gamma=0.8$ & 31.19 & 95.96 & 81.87 & 63.77 & 89.25 & 58.90 & 77.95 \\
LoRA + MixTraining(1:1) & 34.52 & 91.32 & 81.87 & 70.61 & 88.07 & 60.74 & 78.52 \\
MemSFT & 42.92 & 96.48 & 82.99 & 85.40 & 90.49 & 62.77 & 83.62 \\
\bottomrule
\end{tabular}
}
\caption{Baseline results for mitigating forgetting on Qwen3-14B Biology-Instructions. MixTraining uses additional general instruction replay data, while Wise-FT interpolates between the original backbone and the Biology-Instructions LoRA model by scaling the LoRA update. All scores are reported on a 0--100 scale.}
\label{tab:appendix-antiforgetting}
\end{table}

\section{BM25 Retrieval Baseline}
\label{sec:appendix-rag}

We evaluate a direct retrieval-augmented generation baseline on Biology-Instructions to test whether prompting the backbone with retrieved training examples is sufficient for domain specialization \citep{robertson2009probabilistic,lewis2020retrieval}. The base model is Qwen3-14B. We build a single global BM25 index over the same 500K Biology-Instructions training subset and use the original Biology-Instructions evaluation prompt as the query. For each evaluation example, the prompt prepends the retrieved training examples and then asks the model to answer the original problem. We evaluate top-$k$ values of 1, 3, 5, 10, 15, 20, and 50, and report the full per-task Biology-Instructions results in Table~\ref{tab:appendix-rag-full}.

\begin{table}[H]
\centering
\scriptsize
\setlength{\tabcolsep}{2.3pt}
\resizebox{\textwidth}{!}{%
\begin{tabular}{llrrrrrrrr}
\toprule
Task & Metric & Qwen3-14B & Top-1 & Top-3 & Top-5 & Top-10 & Top-15 & Top-20 & Top-50 \\
\midrule
DNA-cpd & MCC & -0.69 & 2.09 & -3.47 & -2.70 & 7.51 & 11.92 & 12.01 & 16.31 \\
DNA-emp & MCC & 0.20 & 1.19 & 3.86 & 2.59 & 1.63 & 3.89 & -2.19 & 4.01 \\
DNA-pd & MCC & 0.22 & -1.65 & -2.48 & -5.86 & -4.46 & 1.00 & -2.40 & -3.28 \\
DNA-tf-h & MCC & -0.72 & -1.39 & 4.11 & 3.90 & 4.61 & 1.14 & 1.85 & 0.44 \\
DNA-tf-m & MCC & 0.28 & 5.51 & 3.82 & 1.82 & 2.54 & 3.93 & 0.05 & 2.88 \\
Multi\_sequence-antibody\_antigen & MCC & -6.04 & 2.17 & -0.56 & -1.39 & -1.28 & 1.50 & 1.51 & 3.18 \\
Multi\_sequence-promoter\_enhancer\_interaction & MCC & -0.64 & 3.44 & -0.81 & -0.72 & -3.51 & -5.68 & -0.17 & 0.58 \\
Multi\_sequence-rna\_protein\_interaction & MCC & -2.89 & 9.22 & 21.99 & 22.35 & 22.38 & 23.99 & 22.01 & 12.92 \\
DNA-enhancer\_activity & PCC & 0.15 & -3.04 & -7.34 & -3.91 & -5.05 & -5.22 & -3.78 & -6.63 \\
RNA-CRISPROnTarget & Spearman & 1.06 & 2.75 & 10.74 & 5.22 & -3.05 & 6.86 & 8.75 & 5.82 \\
Protein-Fluorescence & Spearman & -6.09 & -2.77 & 2.72 & 1.51 & 8.40 & 5.62 & 12.23 & 12.76 \\
Protein-Stability & Spearman & 1.18 & 0.22 & 6.62 & 5.57 & 5.47 & 3.34 & 7.99 & 19.43 \\
Protein-Thermostability & Spearman & -2.36 & -3.81 & -5.33 & -2.13 & -3.42 & -3.81 & -0.05 & 7.68 \\
RNA-Isoform & $R^2$ & 0.17 & 0.02 & 0.00 & 0.10 & 0.21 & 1.33 & 1.67 & 2.18 \\
RNA-MeanRibosomeLoading & $R^2$ & 0.74 & 0.08 & 0.11 & 0.01 & 1.01 & 0.38 & 0.24 & 0.34 \\
RNA-ProgrammableRNASwitches & $R^2$ & 0.09 & 0.00 & 0.16 & 0.20 & 0.04 & 0.20 & 0.26 & 0.98 \\
RNA-Modification & AUC & 50.67 & 49.99 & 50.18 & 49.58 & 49.90 & 49.72 & 49.80 & 49.62 \\
Protein-Solubility & Acc. & 51.50 & 50.30 & 51.60 & 53.00 & 55.30 & 56.70 & 58.70 & 58.40 \\
RNA-NoncodingRNAFamily & Acc. & 15.90 & 12.90 & 11.40 & 11.00 & 10.60 & 12.90 & 13.20 & 12.50 \\
Protein-FunctionEC & Score & 2.93 & 3.12 & 2.88 & 3.20 & 3.23 & 3.62 & 4.53 & 5.09 \\
Multi\_sequence-sirnaEfficiency & Mixed & 33.85 & 38.10 & 42.18 & 45.52 & 46.35 & 48.98 & 48.22 & 51.57 \\
\midrule
Biology-Instructions AVG & -- & 6.64 & 8.02 & 9.16 & 8.99 & 9.45 & 10.59 & 11.16 & 12.23 \\
\bottomrule
\end{tabular}
}
\caption{Full BM25 RAG results on Biology-Instructions with Qwen3-14B. All values are OpenCompass summary metrics. Top-50 uses shorter retrieved-example truncation to keep the prompt within the context budget.}
\label{tab:appendix-rag-full}
\end{table}

\section{Adaptation FLOPs Accounting}
\label{sec:appendix-flops}

This section details the accounting behind Table~\ref{tab:adaptation-flops}. The Biology-Instructions specialization corpus contains $T=90{,}019{,}678$ post-tokenization input tokens for one training epoch. For full SFT, we estimate $6P_{\mathrm{act}}T$ model FLOPs, covering one forward pass and the corresponding backward computation. For LoRA, we use the optimistic lower-bound estimate $4P_{\mathrm{act}}T$, which assumes that freezing the backbone removes the parameter-gradient contribution while retaining the backbone forward and activation-gradient computations. The dense Qwen3-8B, Qwen3-14B, and Qwen3-32B models use their nominal parameter counts, while Qwen3-235B-A22B uses 22B activated parameters per token.

For standalone Qwen3-235B-A22B adaptation, these estimates give 11.88 EFLOPs for SFT and 7.92 EFLOPs for LoRA. Across Qwen3-8B/14B/32B, the summed active parameter count is 54B, giving 29.17 EFLOPs for SFT and 19.44 EFLOPs for LoRA. The corresponding four-backbone totals are therefore 41.05 and 27.37 EFLOPs.

MemSFT separates shared domain-level work from backbone-specific routing. The model-FLOPs accounting for KNN supervision includes a Qwen3-8B forward pass to construct the Biology-Instructions datastore, estimated as $2(8\mathrm{B})T=1.44$ EFLOPs. It does not include FAISS index construction or the subsequent neighbor search and KNN-target materialization, since these system operations are not captured by standard Transformer model-FLOPs accounting. Training the shared 8B memory for one epoch contributes $6(8\mathrm{B})T=4.32$ EFLOPs. Router training uses $T_r=16{,}075{,}270$ tokens, comprising 15,484,539 tokens from the shared general dataset and 590,731 tokens from the BioIns domain subsets. Because both LM branches are frozen, we estimate each backbone-specific router as $2(P_{\mathrm{act}}^{\mathrm{backbone}}+8\mathrm{B})T_r$, corresponding to one forward pass through the backbone and memory; the router's own compute is negligible. The routers for Qwen3-8B/14B/32B jointly contribute 2.51 EFLOPs, while the Qwen3-235B-A22B router contributes 0.96 EFLOPs. Thus, standalone Qwen3-235B-A22B specialization includes 1.44 EFLOPs for datastore construction, 4.32 EFLOPs for memory training, and 0.96 EFLOPs for router training, totaling 6.73 EFLOPs. Reusing the same datastore and memory across all four backbones increases only the router component to 3.47 EFLOPs, yielding a total of 9.23 EFLOPs. These components are summarized in Table~\ref{tab:adaptation-flops}.

To quantify the system operations omitted from the analytical model-FLOPs total, we additionally profile a representative BioIns specialization pipeline with a Qwen3-14B backbone on A800-80GB GPUs. The 8B memory stage is a profiling run that uses the same data and optimization settings as the main experiment. Table~\ref{tab:appendix-flops-wallclock} reports the measured wall-clock and allocation footprint.

\begin{table}[H]
\centering
\footnotesize
\setlength{\tabcolsep}{5pt}
\begin{tabular}{lcccc}
\toprule
Stage & GPUs & Wall-clock & Allocated GPU-hours & Share (\%) \\
\midrule
Datastore construction & 8 & 00:24:25 & 3.26 & 3.57 \\
FAISS index construction & 8 & 00:06:32 & 0.87 & \cellcolor{tablegreen}\textbf{0.95} \\
KNN-target materialization & 8 & 00:05:31 & 0.74 & \cellcolor{tablegreen}\textbf{0.81} \\
8B memory training & 8 & 10:35:11 & 84.69 & 92.77 \\
Qwen3-14B router training & 1 & 01:44:05 & 1.73 & 1.90 \\
\midrule
Total & -- & 12:55:44 & 91.29 & 100.00 \\
\bottomrule
\end{tabular}
\caption{Measured wall-clock and allocation footprint of a representative BioIns specialization pipeline with a Qwen3-14B backbone on A800-80GB GPUs. The 8B memory stage uses the same data and optimization settings as the main experiment. Allocated GPU-hours are computed as the number of allocated GPUs multiplied by wall-clock time.}
\label{tab:appendix-flops-wallclock}
\end{table}

FAISS index construction and KNN-target materialization together take 12 minutes and 3 seconds and account for 1.61 allocated GPU-hours. Relative to the measured pipeline from datastore construction through Qwen3-14B router training, these operations represent 1.55\% of the cumulative wall-clock time and 1.76\% of the allocated GPU-hours. Although they are not included in the analytical model-FLOPs total, their measured overhead is small relative to memory and router training and does not materially alter the overall resource profile in this representative run.

\end{document}